%% file: main.tex
\documentclass[10pt]{article} 
\usepackage[preprint]{tmlr}

\input{math_commands.tex}

\usepackage{url}
\usepackage[table]{xcolor}
\usepackage{booktabs}
\usepackage{mathtools}
\usepackage[most]{tcolorbox}
\usepackage{makecell}
\usepackage{array}

\newcolumntype{P}[1]{>{\raggedright\arraybackslash}p{#1}}

\newtcolorbox[auto counter]{takeaway}[1][]{
  enhanced,
  colframe=black,
  colback=blue!3!white,
  boxrule=0.4mm,
  arc=1.5mm,
  fontupper=\small,
  boxsep=0.5mm,
  left=2mm,
  right=2mm,
  top=4mm,
  bottom=1mm,
  before skip=14pt,
  after skip=8pt,
  title=Takeaway~\thetcbcounter,
  fonttitle=\bfseries\small,
  coltitle=white,
  attach boxed title to top left={xshift=4mm, yshift=-\tcboxedtitleheight/2},
  boxed title style={
    colback=black,
    colframe=black,
    arc=1mm,
    boxrule=0pt,
    top=1mm,
    bottom=1mm,
    left=2mm,
    right=2mm
  },
  #1
}

\title{Sparse Autoencoders Encode Both Concepts and Functions: The Downstream Geometry of Feature Effects}

\author{\name Phu Gia Hoang\thanks{These two authors contributed equally to this work.} 
\email phu.hoang@tu-darmstadt.de \\
       \addr UKP Lab, Technical University of Darmstadt
       \AND
       \name Anwoy Chatterjee\footnotemark[1]
       \email anwoychatterjee@gmail.com \\
       \addr Department of Electrical Engineering\\
       Indian Institute of Technology Delhi, India
       \AND
       \name Tanmoy Chakraborty \email tanchak@iitd.ac.in \\
       \addr Department of Electrical Engineering\\
       Indian Institute of Technology Delhi, India
       \AND
       \name Iryna Gurevych  \email iryna.gurevych@tu-darmstadt.de \\
       \addr UKP Lab, Technical University of Darmstadt
       \AND
       \name Subhabrata Dutta\thanks{Corresponding author.}
       \email subhabrata.dutta@tu-darmstadt.de \\
       \addr UKP Lab, Technical University of Darmstadt
       }

\def\month{MM}  
\def\year{YYYY} 
\def\openreview{\url{https://openreview.net/forum?id=XXXX}} 

\begin{document}

\maketitle

\begin{abstract}
\input{sections/abstract}

\end{abstract}

\input{sections/intro}
\input{sections/related_work}
\input{sections/new_experimental_setup}
\input{sections/feature_isolation}
\input{sections/fega}
\input{sections/analysis}
\input{sections/conclusion}
\input{sections/limitations}
\input{sections/acknowledgement}


\bibliography{main}
\bibliographystyle{tmlr}

\appendix
\input{sections/appendix/tasks}
\input{sections/appendix/metric_derivation}

\end{document}

%% file: math_commands.tex
\usepackage{amsmath,amsfonts,bm}

\def\eqref#1{equation~\ref{#1}}

\def\1{\bm{1}}

\DeclareMathAlphabet{\mathsfit}{\encodingdefault}{\sfdefault}{m}{sl}
\SetMathAlphabet{\mathsfit}{bold}{\encodingdefault}{\sfdefault}{bx}{n}



%% file: sections/abstract.tex
The wide-scale use of sparse autoencoders (SAEs) as interpretability tools is limited by inconsistent links between SAE features and model behavior. Features with clear activation descriptions may have weak or unexpected causal effects; steering can vary across prompts or oppose the intended direction; and activation-based feature selection can miss features that produce the desired output change. Prior work has studied feature geometry inside the model, where features are computed. We instead study the geometry of changes in model logits caused by feature interventions. We introduce Feature-Effect Geometry Analysis (FEGA), an unsupervised framework that removes the same active SAE feature across contexts and analyzes the resulting cloud of logit changes. Across SAE variants, consistent one-dimensional effects are rare: few features behave like reusable directions. To interpret this variation, we distinguish value-like features, tied to static information such as factual attributes, from pointer-like features, associated with context-dependent operations. Value-like features more often exhibit structured, low-dimensional effects, although these effects typically span several directions. Pointer-like features, by contrast, predominantly exhibit diffuse effects. Our results show that a feature can be interpretable and causally relevant without providing a stable direction for steering.

%% file: sections/intro.tex
\section{Introduction} \label{sec:intro}

Sparse autoencoders (SAEs) map model activations into overcomplete sparse latent spaces whose latents\footnote{In the context of SAEs, we use the terms \textit{latent} and \textit{feature} interchangeably in this paper.} are often more interpretable than individual neurons \citep{bricken2023monosemanticity, DBLP:conf/iclr/HubenCRES24}. In mechanistic interpretability~\citep{JMLR:v26:23-0058}, however, activation evidence alone does not establish causal contribution. A causal claim requires intervening on the feature and measuring the resulting change in the model. Such interventions underlie SAE feature steering \citep{templeton2024scaling, durmus2024steering} and appear in benchmark suites such as SAEBench \citep{DBLP:conf/icml/KarvonenRLTBCLF25}. This raises a direct empirical question of \emph{feature-effect consistency}: how comparable are a feature's downstream effects when that feature is intervened on across contexts?

Existing workflows associate SAE features with concepts or behaviors through highly activating examples, automated descriptions, decoder-based readouts, and benchmark-specific criteria \citep{bricken2023monosemanticity, DBLP:conf/iclr/HubenCRES24, DBLP:conf/icml/KarvonenRLTBCLF25, DBLP:journals/corr/abs-2505-20063}. These signals can identify what a feature appears to represent, but not how modifying it propagates through the rest of the model. Prior work shows that steering effects can vary across inputs and prompt formulations, sometimes even opposing the intended direction \citep{tan2024analysing, braun2025understanding}; SAE interventions can produce behavioral effects not predicted by activation contexts \citep{durmus2024steering}; and activation patterns alone may miss features with the desired output effect \citep{DBLP:journals/corr/abs-2505-20063}. These gaps point to a missing measurement: computable diagnostics of how intervention effects are organized, distinguishing reusable effect geometries, structured low-dimensional effect spaces, and diffuse effect clouds that resist low-rank explanation.

\begin{figure}[t]
    \centering
    \includegraphics[width=1\linewidth]{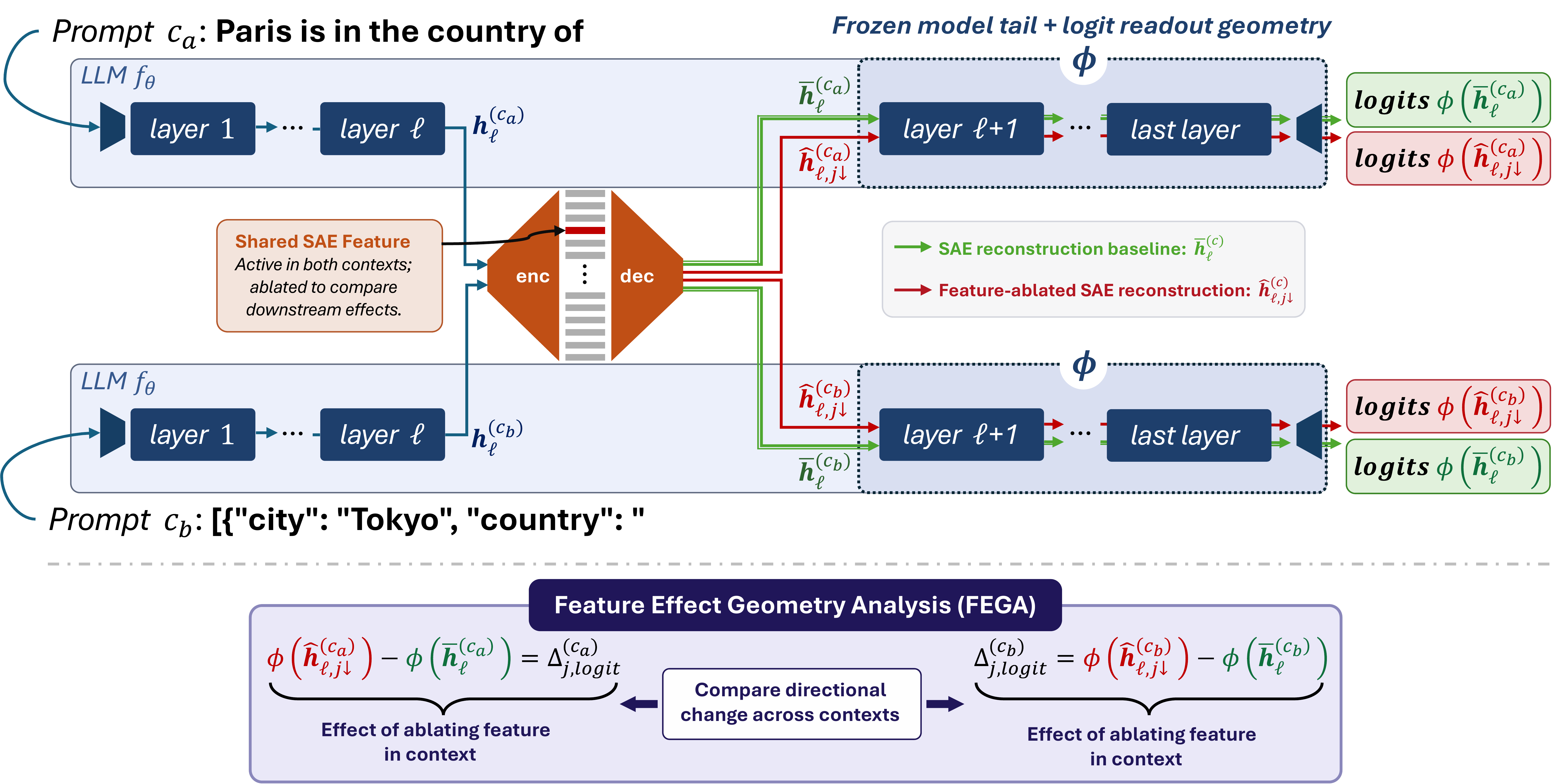}
    \caption{\textbf{SAE feature steering should be evaluated by downstream effect geometry.} Two contexts, $c_a$ and $c_b$, activate the same feature $j$ at layer $\ell$. Green paths show the SAE reconstruction baseline; red paths show the feature-ablated reconstruction. Both are propagated through the frozen model tail and logit readout map $\phi(\cdot)$, producing one downstream logit effect per context, $\Delta_{j,\mathrm{logit}}^{(c_a)}$ and $\Delta_{j,\mathrm{logit}}^{(c_b)}$.}
    \label{fig:teaser}
\end{figure}

What counts as a comparable effect depends on what remains invariant in a feature's role across contexts. We borrow the analogy of values and pointers from programming semantics to distinguish between feature roles. In some cases, a feature is tied to relatively context-agnostic information; we call this kind of role \emph{value-like}. 

For example, a feature associated with the concept of Paris may be relevant whether the prompt asks for the capital of France, discusses European cities, or compares tourist destinations, because its role is tied to the same underlying information. Such a feature might be expected to produce similar downstream effects across these contexts. In other cases, the invariant is not the information itself but the function the feature supports; we call this kind of role \emph{pointer-like}. For example, in a copying or rule-following setting, a feature may support the same operation while acting on values supplied by the context.~\footnote{Existing SAEs report similar patterns on the input side. For example, \url{https://www.neuronpedia.org/gemma-2-9b-it/20-gemmascope-res-131k/78155} demonstrates a feature that activates for every token after a punctuation.} However, prior work does not explore whether such patterns hold for outputs. The key implication is that a feature can make a coherent causal contribution without corresponding to a single shared logit direction, motivating our question of whether its effects exhibit a reproducible geometric structure when the same intervention is repeated across contexts.

We study this distinction with tasks that vary whether the model retrieves stable pretrained information or follows prompt-local operations. On the value-like end, RAVEL city-country evaluates attribute disentanglement, as implemented in SAEBench, where features are selected through entity-attribute interventions \citep{huang-etal-2024-ravel, DBLP:conf/icml/KarvonenRLTBCLF25}. Conversely, synthetic tasks such as Literal Sequence Copying and Word Content rely on randomly sampled tokens to test pattern matching, while PrOntoQA uses fictional predicates and prompt-specified rules to test prompt-local rule completion; these give pointer-like settings that operate without fixed token meanings \citep{niu2025illusion, saparov2023language}. Bridging these settings, Token Translation combines an in-context translation format with pretrained lexical associations, serving as a mixed setting \citep{niu2025illusion}.

To evaluate feature behavior across these tasks, Figure~\ref{fig:teaser} illustrates our core intervention method. For a given context $c$, let $h_\ell^{(c)}$ be the activation at the SAE site, with encoding $z^{(c)}=E(h_\ell^{(c)})$ and reconstruction $\bar h_\ell^{(c)}=D(z^{(c)})$. We define the reconstruction after ablating feature $j$ as $\hat h_{\ell,j\downarrow}^{(c)}=D(z^{(c)}-z_j^{(c)}e_j)$, where $e_j$ is the $j$th SAE basis vector. We use $\bar h_\ell^{(c)}$, rather than the original activation $h_\ell^{(c)}$, as the baseline so that the two paths differ only in whether feature $j$ is retained. Comparing $\hat h_{\ell,j\downarrow}^{(c)}$ directly with $h_\ell^{(c)}$ would additionally include SAE reconstruction error, confounding feature removal with the replacement of the original activation by its reconstruction. We run the reconstructed baseline and the feature-ablated reconstruction through the frozen model tail and compare their downstream logit readouts:

$$
\Delta_{j,\mathrm{logit}}^{(c)} = \phi(\hat h_{\ell,j\downarrow}^{(c)}) - \phi(\bar h_\ell^{(c)}).
$$

The set $\mathcal E_{j,\mathrm{logit}}={\Delta_{j,\mathrm{logit}}^{(c)} : c \in \mathcal C_j^{\mathrm{valid}}}$ is the feature's downstream logit-effect cloud, with the valid-context set defined formally in Section~\ref{sec:fega_setup}. The local decoder geometry fixes the immediate reconstruction change caused by removing feature $j$, but it does not determine how that change propagates through later attention blocks, MLPs, residual streams, and final normalization. The intervention-relevant question is, therefore, geometric: \emph{what shape does $\mathcal E_{j,\mathrm{logit}}$ have in the logit-readout space?}

To this end, we introduce \textbf{Feature-Effect Geometry Analysis (FEGA)}, an offline, unsupervised framework for measuring and classifying downstream effect clouds \footnote{Project repository: https://github.com/UKPLab/FEGA}. FEGA treats each feature's intervention effects as a cloud rather than a single steering object. It first tests whether the effects share one dominant readout direction; if not, it asks how they vary across contexts: through magnitude changes, sign flips along an axis, multiple directional modes, low-dimensional spans, or diffuse structure. We apply FEGA to residual-stream SAEs trained on the post-layer-12 residual stream of Gemma-2-2B~\citep{lieberum-etal-2024-gemma} and released in the SAEBench baseline suite~\citep{DBLP:conf/icml/KarvonenRLTBCLF25}. We compare ReLU~\citep{bricken2023monosemanticity}, TopK~\citep{gao2025scaling}, and Matryoshka Batch TopK~\citep{pmlr-v267-bussmann25a} variants of width $65$k across value-like factual-attribute interventions, pointer-like copying and rule-following tasks, and a mixed token-translation setting. 

Across this evaluation spectrum, downstream feature effects are substantially more context-dependent than a single-vector interpretation suggests. Our main findings are:

\begin{itemize}
    \item \textbf{Feature effects rarely collapse to a single direction (Section~\ref{sec:geometric_analysis}).} Across architectures and feature populations, most downstream effect clouds do not reduce to a context-invariant logit vector, suggesting that simple linear steering is the exception rather than the rule.
    
    \item \textbf{Pointer-like behavior is associated with a compact, partially shared set of SAE features (Sections~\ref{subsec:architecture_impact} and
    \ref{subsec:feature_overlap}).} Across copying, binding, rule application, and translation tasks, most isolated features are task-specific, but a small number of features recur across multiple tasks. This cross-task reuse is more pronounced in TopK and Matryoshka Batch TopK than in ReLU SAEs, while the pointer-like sets remain disjoint from the RAVEL value-like feature sets.
   
    \item \textbf{Pointer-like features produce predominantly diffuse downstream effects
    (Section~\ref{subsec:geometry_pointer}).}
    Despite their recurrence across successful task executions, features associated with prompt-local operations rarely act as reusable logit directions. Their effects scatter across output space, consistent with the output target changing with the value being copied, bound, or translated.
 
    \item \textbf{Value-like features have more structured downstream effects (Section~\ref{subsec:geometry_value}).} Compared with pointer-like effects, factual-attribute effects more often show low-dimensional organization. This structure typically spans multiple output directions rather than collapsing to a single steering vector.
\end{itemize}

These findings challenge a common assumption in feature-level interpretability: that a latent with a clear description and measurable causal effect can serve as a reliable explanatory or control variable. Prior automated-interpretability work has emphasized quantitative scoring of explanations, including human comparisons, while showing that activation descriptions can miss downstream functions and out-of-distribution behavior \citep{bills2023language}. FEGA extends this perspective from activation description to intervention effect by asking whether a feature's causal effects form stable geometry across contexts. Our results suggest this stability is often limited: an SAE feature can be interpretable and causally important while lacking a consistent logit-space direction. Pointer-like features show strong context dependence and predominantly diffuse downstream effects. Value-like features show low-dimensional organization more frequently, but their structured effects typically span several output directions rather than forming a single steering vector. Together, these results suggest that feature-level interpretability should distinguish what a feature detects, whether it matters causally, and how its effects vary across contexts. FEGA provides this missing effect-side audit.

%% file: sections/related_work.tex
\section{Related Work} \label{sec:related_work}

In this section, we restrict the discussion to the literature related to geometry, steering, and semantic multiplicity of SAE features. For a comprehensive overview of SAEs in general, refer to \citet{DBLP:journals/corr/abs-2503-05613} and \citet{DBLP:journals/tmlr/SharkeyCBLWBGHOBBGCNRWS25}.

\indent\textbf{\textit{Geometry of SAE Features.}} Uncovering the geometric properties of SAE features has garnered interest recently. One of the foundational works in this direction, by \citet{NEURIPS2025_110d919b}, investigates the structural assumptions associated with SAE architectures and how concepts are encoded in model representations. \citet{DBLP:journals/entropy/LiMBEST25} show that SAEs organize concepts differently across scales, while SAEs have also been used to uncover belief-state geometries~\citep{DBLP:journals/corr/abs-2604-02685}. \citet{DBLP:journals/corr/abs-2604-28119} show that SAEs can capture manifolds either globally, through groups of atoms whose span contains the manifold, or locally, through features that tile restricted regions. Our work, though related to this line of research in terms of geometric interests, focuses on a different object altogether: effects of SAE features in the downstream logit space.

\indent\textbf{\textit{Steering Model Behavior with SAE Features.}} Prior studies have identified inconsistencies in the role of SAE features in predicting downstream behaviors. \citet{DBLP:conf/eacl/LiSBL26} observe that even tiny adversarial input perturbations can manipulate concept representations in SAEs. \citet{DBLP:journals/corr/abs-2602-14111} investigate SAEs with frozen (randomly initialized) encoders and decoders; they demonstrate causal editing and sparse probing performance comparable to fully-trained SAEs, consequently questioning the ability of SAEs to learn meaningful features. \citet{DBLP:conf/icml/WuAG00JMP25} highlight that SAE features are unreliable for steering the model behavior, often outperformed by simpler baselines. On the other hand, \citet{DBLP:journals/corr/abs-2505-20063} show that the steerability is dependent on feature types; they identify two distinct types of SAE features: input features, which mainly capture patterns in the model's input, and output features, which have a human-understandable effect on the model's output. While most prior attempts in this broad direction uncover many interesting success/failure modes of SAE features, they do not characterize the geometric properties of these features that cause such peculiarities.

\indent\textbf{\textit{Polysemanticity of SAE Features.}} A representation (e.g., a feature or a neuron) is polysemantic when it is associated with multiple unrelated concepts. The original goal of SAEs in mechanistic interpretability is to extract monosemantic features from polysemantic representations of the model~\citep{bricken2023monosemanticity}. \citet{DBLP:conf/iclr/MinegishiFIM25} propose an evaluation strategy based on polysemous words to determine how good an SAE is in extracting monosemantic features, ultimately highlighting the limitations of most prevalent SAE architectures. The theoretical analysis laid out by \citet{cui2026on} reveals that SAEs fail to fully recover the ground-truth monosemantic features unless the features are extremely sparse. \citet{DBLP:journals/corr/abs-2605-14694} shed light on this problem from a rate-distortion perspective, showing that forcing the SAE to learn monosemantic features comes at the cost of higher reconstruction error and lower sparsity, ultimately leading to a trade-off. While the degree of semantic multiplicity is fundamental to the {\em what is represented by a feature} in the concept level, it does not answer the subsequent question of {\em what will happen if a certain feature is strengthened/weakened}.

Our work, to the best of our knowledge, is the first of its kind to characterize the geometric properties of SAE features in the effect space, tying together the missing pieces of feature geometry, steering and semantic multiplicity. 

%% file: sections/new_experimental_setup.tex
\section{Experimental Setup}
\label{sec:experimental_setup}
This section defines the experimental setting for our downstream geometry analysis. We first specify the base model and SAE variants, then introduce the task spectrum used to sample value-like, pointer-like, and mixed feature populations. Feature isolation itself is described separately in Section~\ref{sec:feature_isolation}.

\subsection{Model and SAE Configurations}

We perform all experiments using Gemma-2-2B \citep{DBLP:journals/corr/abs-2408-00118}, analyzing SAEs applied to the post-layer-12 residual stream. To check that our conclusions are not artifacts of a single sparsity mechanism, we evaluate three SAE variants from the SAEBench Gemma-2-2B suite \citep{DBLP:conf/icml/KarvonenRLTBCLF25}: \textit{ReLU}, \textit{TopK}, and \textit{Matryoshka Batch TopK}. Each SAE has a width (i.e., dictionary size) of $2^{16}$, i.e., $\sim 65$k.

\subsection{The Task Spectrum}
\label{subsec:tasks}

We use five tasks spanning synthetic in-context operations to factual attribute retrieval. For each of the four in-context learning (ICL) tasks, we construct $50{,}000$ examples balanced across $1{,}000$ prompt families, giving 50 retained queries per family. We retain only examples with single-token targets for which the unmodified Gemma-2-2B model predicts the correct target as its first answer token. A \textit{prompt family} denotes a fixed template, demonstration context, or prompt-local mapping, while a \textit{query} denotes one completion instance generated from that family. This sampling design helps distinguish features associated with reusable prompt-local operations from features associated with incidental tokens in individual prompts.

\indent\textbf{\textit{Literal Sequence Copying (LSC).}} LSC \citep{niu2025illusion} is the purely synthetic endpoint of our spectrum. Each prompt family samples a random token pattern $P$, inserts a target token $T$ after its first occurrence, and repeats $P$ later. The query asks the model to output the token that followed the earlier occurrence of the same pattern. For example: 

\begin{quote}\small 
\texttt{attorney impair georgia all berry enlarged takes any dispute ...}\\
\texttt{return attorney impair georgia all berry} 
\end{quote} 

has answer \texttt{enlarged}. Since sampled words are arbitrary, solving LSC requires a minimal pointer-like operation: \emph{find the previous occurrence of this pattern and copy what came next}.

\indent\textbf{\textit{Word Content (WC).}} Word Content \citep{niu2025illusion} keeps a synthetic vocabulary but adds a classification structure. Each prompt family assigns labels to sets of trigger tokens, and each query contains one trigger set plus distractors. For example, a family may define:
\[ 
\texttt{produce} + \texttt{smell} \rightarrow \texttt{find}, 
\qquad 
\texttt{names} + \texttt{drop} \rightarrow \texttt{our}. 
\] 
A query such as:
\begin{quote}\small 
\texttt{agreed age miscellaneous reduced might meantime force smell produce} 
\end{quote} 
has answer \texttt{find}. WC, therefore, tests the pointer-like operation of \emph{identifying prompt-local evidence and emitting the associated label, without relying on semantic shortcuts.}

\indent\textbf{\textit{PrOntoQA.}} We adapt PrOntoQA \citep{saparov2023language} to construct prompt-local rule-completion tasks with fictional predicates and newly generated entity names. Each prompt family provides demonstrations associating source predicates with target predicates. A query introduces a new entity with one of the demonstrated source predicates and asks the model to complete the corresponding rule:

\begin{quote}\small 
\texttt{Q: Sam is a storpist. Every storpist is a stopin. Sam is a}\\ 
\texttt{A: stopin}\\ \texttt{...}\\ 
\texttt{Q: Siomdu is a storpist. Every storpist is a}\\ 
\texttt{A:} 
\end{quote} 

with answer \texttt{stopin}. Because predicates such as \texttt{storpist} and \texttt{stopin} are fictional, success depends on retrieving the demonstrated source-to-target predicate binding from context rather than relying on world knowledge. We therefore treat PrOntoQA as a pointer-like task.

\indent\textbf{\textit{Token Translation (TT).}} Token Translation \citep{niu2025illusion} combines prompt-local task specification with pretrained lexical knowledge. Each prompt family fixes a source and target language and gives several translation demonstrations. A query then asks for a new translation in the same format: 

\begin{quote}\small 
\texttt{Translate English words into German.}\\ 
\texttt{English: chair}\\ 
\texttt{German: Stuhl}\\ 
\texttt{English: ring}\\ 
\texttt{German: Ring}\\ 
\texttt{...}\\ 
\texttt{English: apple}\\ 
\texttt{German:} 
\end{quote} 

with answer \texttt{Apfel}. TT is not purely synthetic: the model must follow the translation schema established by the prompt, while the final answer depends on pretrained lexical associations. We therefore treat TT as a hybrid task combining pointer-like schema following with value-like lexical retrieval.

\indent\textbf{\textit{RAVEL.}} RAVEL (Resolving Attribute-Value Entanglements in Language Models), proposed by \citet{huang-etal-2024-ravel}, forms the semantic endpoint of our spectrum. Unlike the ICL tasks, RAVEL does not provide a prompt-local mapping. Its instances are organized by entity, queried attribute, and natural-language prompt template. We analyze the city entity class, treating \textsc{Country} as the target attribute and attributes such as \textsc{Language} and \textsc{Continent} as non-target controls. A base query such as:

\begin{quote}\small 
\texttt{The city of Paris is located in the country of} 
\end{quote} 
has expected completion \texttt{France}, and a source query such as: 
\begin{quote}\small 
\texttt{The city of Tokyo is located in the country of} 
\end{quote} 
has expected completion \texttt{Japan}.

RAVEL asks whether transferring the relevant internal representation from the source can make the model answer \texttt{Japan} for the Paris query while preserving the base entity's non-target attributes, such as its \textsc{Language} or \textsc{Continent}. Thus, factual recall provides the underlying task, while the intervention evaluates whether the target attribute can be causally isolated from other factual attributes. We use the resulting country-associated features as the value-like endpoint of our spectrum because their outputs depend on pretrained factual associations rather than a prompt-local mapping.

%% file: sections/feature_isolation.tex
\section{Isolating Value-Like and Pointer-Like Features}
\label{sec:feature_isolation}

\begin{table}[t]
    \centering
    \footnotesize
    \begin{tabular}{lcccc}
        \toprule
        \textbf{Feature Source} & \textbf{Regime} & \textbf{ReLU SAE} & \textbf{TopK SAE} & \makecell{\textbf{Matryoshka}\\\textbf{Batch TopK SAE}} \\
        \arrayrulecolor{black}
        \midrule
        LSC & Pointer-Like & 5 & 5 & 4 \\
        \arrayrulecolor{gray!25}
        \hline
        WC & Pointer-Like & 25 & 9 & 14 \\
        \hline
        PrOntoQA & Pointer-Like & 78 & 22 & 20 \\
        \hline
        TT & Hybrid & 27 & 14 & 16 \\
        \hline
        RAVEL (City-Country) & Value-Like & 12,900 & 2,828 & 3,654 \\
        \arrayrulecolor{black}
        \bottomrule
    \end{tabular}
\caption{\textbf{Identified feature sets.}
Number of features selected for each task across Gemma-2-2B SAE variants of width $65$k.}
    \label{tab:feature_counts}
\end{table}

The task spectrum in Section~\ref{subsec:tasks} ranges from prompt-local operations to factual retrieval. We use this spectrum to construct and compare two task-conditioned feature populations: \emph{value-like} features associated with pretrained factual attributes, and \emph{pointer-like} features recurrently associated with prompt-local operations.

\subsection{Identifying Value-Like Features}
\label{subsec:identifying_values}

To isolate value-like features, we use the RAVEL attribute-disentanglement setup through SAEBench. SAEBench applies differential binary masking \citep{DBLP:journals/corr/abs-2409-04478} over SAE latents to find the smallest feature subset needed for an interchange intervention. Because these features are selected to control localized semantic attributes, they are natural value-like candidates. Starting from 5{,}000 base records, differential binary masking selects 12{,}900 ReLU, 2{,}828 TopK, and 3{,}654 Matryoshka Batch TopK features. Requiring at least eight active contexts leaves 7{,}715, 1{,}167, and 1{,}750 features, respectively, for downstream geometric analysis.

\subsection{Identifying Pointer-Like Features}
\label{subsec:identifying_pointers}

To identify pointer-like candidates, we analyze the model-correct examples from the four ICL tasks described in Section~\ref{sec:experimental_setup}: LSC, WC, PrOntoQA, and TT. At the final token before prediction, we consider an SAE feature active if its activation is positive. We retain a feature if it is active on at least 90\% of all examples and on at least 90\% of the queries in at least 90\% of prompt families. These thresholds select features that are active across many prompts and outputs, rather than only for specific words, queries, or prompt families. This procedure selects between $4$ and $78$ features for each task--architecture pair; Table~\ref{tab:feature_counts} reports the full counts.

\begin{figure}[t]
    \centering
     \includegraphics[width=\linewidth]{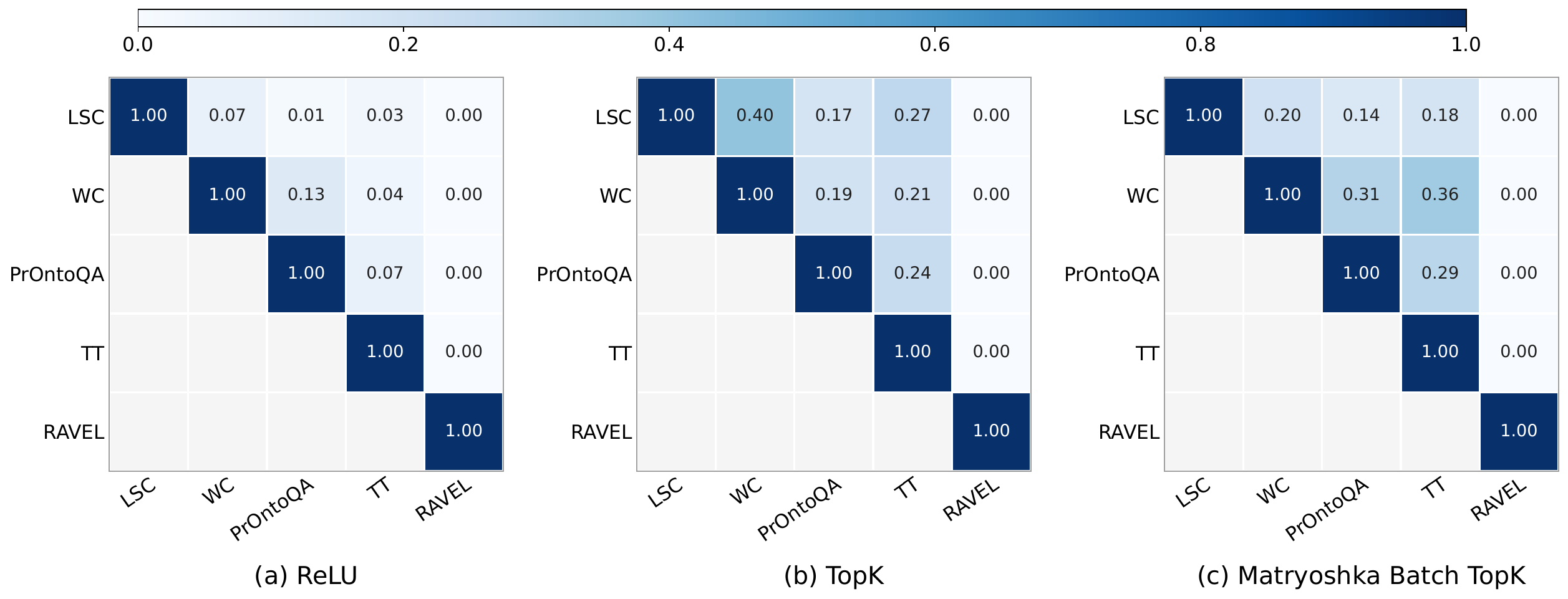}
\caption{\textbf{Cross-task feature overlap.}
Intersection over Union (IoU) of the selected feature sets. Several features recur across the ICL tasks, with greater overlap for TopK and Matryoshka Batch TopK than for ReLU. None of the ICL-selected features overlaps with the corresponding RAVEL feature set.}
    \label{fig:feature_overlap}
\end{figure}

\subsection{Feature Discovery Across SAE Architectures}
\label{subsec:architecture_impact}

Table~\ref{tab:feature_counts} shows how selected set sizes vary across tasks and SAE architectures. No architecture consistently yields the largest set: ReLU identifies the most candidates for WC, PrOntoQA, and TT, whereas all three architectures produce similarly small sets for LSC.

The number of selected features is not, by itself, a measure of how well an architecture captures task-related computations. Feature activation patterns and sparsity differ across architectures, making the raw counts difficult to compare directly. We therefore examine cross-task overlap to determine whether features are shared across tasks and use ablation to measure their contribution to task performance.

\begin{takeaway}
\textbf{Pointer-like recurrence is concentrated in small feature sets across architectures.}
The number of selected candidates varies across tasks and SAE variants, but no architecture uniformly dominates functional feature discovery.
\end{takeaway}




\subsection{Shared Features Across In-Context Tasks}
\label{subsec:feature_overlap}

If particular SAE features contribute to general ICL and induction-like behavior, they should recur across tasks despite differences in vocabulary, format, and output structure. We therefore compare the selected feature sets across the four in-context tasks. Figure~\ref{fig:feature_overlap} shows clear evidence of such reuse. For TopK, the largest overlap is between LSC and WC, with an IoU of $0.40$. For Matryoshka Batch TopK, WC overlaps with TT at $0.36$ and with PrOntoQA at $0.31$. ReLU exhibits less cross-task reuse, with a maximum IoU of $0.13$ between WC and PrOntoQA.

Because the selected sets are small, these overlaps point to a compact group of features shared across multiple in-context tasks. This shared component may support computations common to ICL and induction-like behavior, while the remaining features may reflect task-specific demands such as input structure, formatting, or output type. Having identified this shared and task-specific structure, we next use ablation in Section~\ref{subsec:causal_ablation} to test whether the selected feature sets causally affect task performance.

For all three SAE architectures, no features selected from an ICL task appears in the RAVEL City--Country feature set. Under our selection procedures, features associated with prompt-local operations are therefore distinct from those associated with factual attribute disentanglement. This result applies to the selected candidate sets and does not imply that the entire SAE dictionary separates cleanly into pointer-like and value-like regions.

\begin{takeaway}
\textbf{Some SAE features recur across distinct in-context tasks.}
Their reuse points to a compact shared component underlying ICL and induction-like behavior, while the selected ICL features remain distinct from the RAVEL-selected factual-attribute features.
\end{takeaway}



\begin{table*}[t]
    \centering
    \footnotesize
    \setlength{\tabcolsep}{8pt}
    \renewcommand{\arraystretch}{1.12}
    \begin{tabular}{lrcrcr}
        \toprule
        \textbf{Task} & \textbf{$k$} & \makecell{\textbf{Targeted Ablation}\\\textbf{Accuracy (in \%)}} & \textbf{$p_{\mathrm{target}}$} & \makecell{\textbf{Random Control}\\\textbf{Accuracy (in \%)}} & \textbf{$p_{\mathrm{random}}$} \\
        \arrayrulecolor{black}
        \hline
        \rowcolor{gray!15} \multicolumn{6}{c}{\textbf{ReLU SAE}} \\\hline
        LSC & 5 & 95.7 & $<10^{-300}$ & 98.3$\pm$0.4 & $5.3\times10^{-15}$ \\
        \arrayrulecolor{gray!25}
        \hline
        WC & 25 & 76.5 & $<10^{-300}$ & 93.4$\pm$4.0 & $3.0\times10^{-7}$ \\
        \arrayrulecolor{gray!25}
        \hline
        PrOntoQA & 78 & 71.3 & $<10^{-300}$ & 94.4$\pm$1.4 & $1.1\times10^{-13}$ \\
        \arrayrulecolor{gray!25}
        \hline
        TT & 27 & 89.9 & $<10^{-300}$ & 98.7$\pm$0.5 & $1.6\times10^{-10}$ \\
        \arrayrulecolor{black}
        \hline
        \rowcolor{gray!15} \multicolumn{6}{c}{\textbf{TopK SAE}} \\\hline
        LSC & 5 & 65.1 & $<10^{-300}$ & 95.7$\pm$1.1 & $1.7\times10^{-13}$ \\
        \arrayrulecolor{gray!25}
        \hline
        WC & 9 & 18.4 & $<10^{-300}$ & 93.6$\pm$4.0 & $4.7\times10^{-7}$ \\
        \arrayrulecolor{gray!25}
        \hline
        PrOntoQA & 22 & 47.1 & $<10^{-300}$ & 95.7$\pm$1.8 & $1.1\times10^{-9}$ \\
        \arrayrulecolor{gray!25}
        \hline
        TT & 14 & 72.9 & $<10^{-300}$ & 98.2$\pm$0.6 & $9.0\times10^{-12}$ \\
        \arrayrulecolor{black}
        \hline
        \rowcolor{gray!15} \multicolumn{6}{c}{\textbf{Matryoshka Batch TopK SAE}} \\\hline
        LSC & 4 & 64.3 & $<10^{-300}$ & 95.8$\pm$2.1 & $1.6\times10^{-8}$ \\
        \arrayrulecolor{gray!25}
        \hline
        WC & 14 & 19.9 & $<10^{-300}$ & 94.7$\pm$1.0 & $6.8\times10^{-16}$ \\
        \arrayrulecolor{gray!25}
        \hline
        PrOntoQA & 20 & 17.9 & $<10^{-300}$ & 92.1$\pm$4.6 & $1.5\times10^{-7}$ \\
        \arrayrulecolor{gray!25}
        \hline
        TT & 16 & 66.0 & $<10^{-300}$ & 97.5$\pm$2.4 & $9.2\times10^{-5}$ \\
        \arrayrulecolor{black}
        \bottomrule
    \end{tabular}
\caption{\textbf{Causal effects of identified ICL features.}
Accuracy after jointly ablating the $k$ selected features at the final prompt position. Results are evaluated on examples answered correctly by the unablated model. Random controls ablate same-sized feature sets matched by activation prevalence and magnitude; values report mean$\pm$standard deviation across 20 trials. $p_{\mathrm{target}}$ is obtained from a one-sided exact paired McNemar test, and $p_{\mathrm{random}}$ from a one-sided one-sample $t$-test across the random trials.}
    \label{tab:icl_ablation_results}
\end{table*}

\subsection{Causal Contribution via Ablation}
\label{subsec:causal_ablation}

Recurrence and overlap establish an association with prompt-local behavior, but not a causal contribution. We therefore jointly ablate each selected feature set at the final prompt position and measure the resulting task accuracy. The intervention preserves the SAE reconstruction error: we encode the original residual-stream activation, zero the selected SAE coordinates, decode the modified representation, and add the original reconstruction error before patching it back into the model.

For comparison, we run 20 control ablations. Each control removes an equally sized set of other features chosen to have similar activation frequency and average activation strength. As shown in Table~\ref{tab:icl_ablation_results}, targeted ablation lowers accuracy in all 12 task--architecture combinations. The effect ranges from a $4.3$ percentage-point drop for ReLU on LSC to an $82.1$ percentage-point drop for Matryoshka Batch TopK on PrOntoQA. In every case, the targeted drop is larger than the mean drop across the matched random controls. For example, ablating the 9 TopK WC features reduces accuracy from $100\%$ to $18.4\%$ ($p<10^{-300}$), whereas the matched random controls retain an average accuracy of $93.6\%\pm4.0$.

These results show that the selected feature sets make a specific causal contribution to task performance beyond what is expected from ablating similarly active feature sets of the same size. They do not imply that every selected latent is individually necessary.

\begin{takeaway}
\textbf{Selected pointer-like feature sets contribute causally to task performance.}
Jointly ablating them produces consistently larger drops in accuracy compared to same-sized matched random controls.
\end{takeaway}




\subsection{The Need for Downstream Geometric Analysis}
\label{subsec:need_for_geometry}

The selection, overlap, and ablation results establish task-conditioned feature populations with different functional associations. RAVEL identifies features involved in factual attribute disentanglement, while recurrence and ablation identify small feature sets that are reused across prompt-local tasks and causally affect task performance.

These results do not show that a feature's downstream effects are stable across contexts. A feature may activate reliably across copying prompts, yet its ablation may perturb different logit directions depending on the prompt structure or copied token. Its effect cloud may align with one direction, separate into context-dependent regimes, or remain diffuse in vocabulary space. Thus, after grouping features by the behaviors through which they were isolated and establishing their causal relevance, we ask how their downstream effects vary across contexts. This motivates the geometric analysis that follows.

%% file: sections/fega.tex
\section{Feature-Effect Geometry Analysis (FEGA)}
\label{sec:method}

To address the need for downstream geometric analysis established in Section~\ref{subsec:need_for_geometry}, we introduce \textbf{Feature-Effect Geometry Analysis} (FEGA). FEGA studies the downstream readout change produced by ablating an active SAE feature and patching the resulting reconstruction back into the model. Across contexts, these effects form an empirical cloud in logit space. FEGA analyzes this cloud to determine whether a feature acts as one consistent steering direction or instead follows another geometric pattern, summarized by a diagnostic profile and a primary geometry label.

\subsection{Downstream Feature Effects}
\label{sec:fega_setup}

To construct each effect cloud, we define an intervention relative to the SAE reconstruction. We first specify this reconstruction-relative baseline, then formalize feature zeroing, downstream readout effects, valid contexts, and the sense in which downstream effects can depart from the local decoder direction.

\indent\textbf{\textit{Contexts and SAE Reconstruction.}} A context $c$ determines both the prompt and the task-specific analysis position. Let $h_\ell^{(c)}\in\mathbb R^{d_{\mathrm{model}}}$ be the original model activation at the SAE input site in layer $\ell$. The SAE encoder and decoder are $E:\mathbb R^{d_{\mathrm{model}}}\to\mathbb R^{d_{\mathrm{sae}}}$ and $D:\mathbb R^{d_{\mathrm{sae}}}\to\mathbb R^{d_{\mathrm{model}}}$, giving:

$$
z^{(c)} = E(h_\ell^{(c)})\in\mathbb R^{d_{\mathrm{sae}}},
\qquad
\bar h_\ell^{(c)} = D(z^{(c)})\in\mathbb R^{d_{\mathrm{model}}}.
$$

All feature-ablation comparisons below are reconstruction-relative: the baseline path patches $\bar h_\ell^{(c)}$, rather than the original activation $h_\ell^{(c)}$, so that the compared paths differ only in whether feature $j$ is retained and are not confounded by SAE reconstruction error.

\indent\textbf{\textit{Feature-Zeroing Intervention.}}
For feature $j\in\{1,\ldots,d_{\mathrm{sae}}\}$, define the active context set $\mathcal C_j=\{c : z_j^{(c)}>0\}.$ On an active context, feature-zeroing removes only coordinate $j$ from the SAE code:

$$
z_{j\downarrow}^{(c)} = z^{(c)} - z_j^{(c)} e_j,
\qquad
\hat h_{\ell,j\downarrow}^{(c)} = D(z_{j\downarrow}^{(c)}),
$$

where $e_j\in\mathbb R^{d_{\mathrm{sae}}}$ is the $j$th feature basis vector. We then patch $\bar h_\ell^{(c)}$ and $\hat h_{\ell,j\downarrow}^{(c)}$ into the same SAE site and target position in separate forward passes, so the two paths differ only in whether feature $j$ is retained in the reconstruction.

\indent\textbf{\textit{Frozen Tail and Logit Readout.}}
Let $r_\ell^{(c)}:\mathbb R^{d_{\mathrm{model}}}\to\mathbb R^{d_{\mathrm{model}}}$ denote the frozen model tail from the SAE site in layer $\ell$ to the vector presented to the output embedding at the target position, with patched activation $u$ inserted at that site. For Gemma-2-2B, this vector is taken after the model’s final normalization. The superscript $(c)$ records that the surrounding prompt is fixed, while the model weights remain unchanged. With unembedding matrix $W_U\in\mathbb R^{|\mathcal V|\times d_{\mathrm{model}}}$, the corresponding linear logit readout, before the output soft cap, is 

$$
\phi_\ell^{(c)}(u)=W_U r_\ell^{(c)}(u)\in\mathbb R^{|\mathcal V|}.
$$

\indent\textbf{\textit{Removal Effects.}}
The canonical FEGA object is the logit-space removal effect:

$$
\Delta_j^{(c)}=
\phi_\ell^{(c)}(\hat h_{\ell,j\downarrow}^{(c)})-
\phi_\ell^{(c)}(\bar h_\ell^{(c)})
\in\mathbb R^{|\mathcal V|}.
$$

Thus, $\Delta_j^{(c)}$ is always oriented as ablated readout minus reconstruction-baseline readout. A positive coordinate of $\Delta_j^{(c)}$ means that the corresponding token logit increases after feature $j$ is removed; the opposite direction, $-\Delta_j^{(c)}$, describes what the retained feature contributed before removal. Appendix~\ref{app:removal_contribution_sign_invariance} gives the sign-orientation details.

Rather than storing a vocabulary-dimensional vector for every context, we retain the corresponding removal effect at the input to the output embedding,

$$
\delta_j^{(c)}=
r_\ell^{(c)}(\hat h_{\ell,j\downarrow}^{(c)})-
r_\ell^{(c)}(\bar h_\ell^{(c)})
\in\mathbb R^{d_{\mathrm{model}}}.
$$

Its induced linear-logit effect is $\Delta_j^{(c)}=W_U\delta_j^{(c)}$. Thus, FEGA measures geometry in linear-logit space while computing its magnitudes and inner products from the smaller stored vectors.

\indent\textbf{\textit{Effect Clouds and Valid Contexts.}}
For feature $j$, FEGA analyzes the finite empirical cloud of valid logit removal effects $\mathcal E_j=\{\Delta_j^{(c)} : c\in\mathcal C_j^{\mathrm{valid}}\}.$ We retain a context only when the two readout vectors and their difference are finite and the induced linear-logit effect is larger than numerical zero. Specifically,

$$
\mathcal C_j^{\mathrm{valid}}=
\left\{
c\in\mathcal C_j :
\Delta_j^{(c)} \text{ is finite and non-zero}
\right\}.
$$

We validate this implementation on a held-out subset where both $\delta_j^{(c)}$ and explicit $\Delta_j^{(c)}$ are materialized. The cloud $\mathcal E_j$ is not the full distribution of all possible contexts; it is the sampled set of valid effects under the chosen data set, feature-selection rule, target position, reconstruction baseline, and removal-effect orientation.

\indent\textbf{\textit{Why Downstream Effects Are Needed?}}
For a linear SAE decoder with decoder matrix
$W_D\in\mathbb R^{d_{\mathrm{model}}\times d_{\mathrm{sae}}}$, zeroing one feature gives the local reconstruction perturbation:
$$
\hat h_{\ell,j\downarrow}^{(c)}-\bar h_\ell^{(c)}=
D(z^{(c)}-z_j^{(c)}e_j)-D(z^{(c)})=
-z_j^{(c)}W_{D,:,j}.
$$

This local decoder-removal term has a fixed direction scaled by the feature activation. The measured effect $\Delta_j^{(c)}$, however, is defined after the patched activation passes through later attention blocks, MLPs, residual additions, final normalization, and the unembedding. These downstream computations can make the logit removal effect context-dependent even when the local decoder-removal direction is fixed.

The central question is, therefore, geometric: \emph{does the sampled logit-effect cloud $\mathcal E_j$ behave like one stable directed removal effect, or does it vary across contexts in a structured way?} Later in this section, FEGA equips these effects with logit-space geometry, separates magnitude from direction, and tests whether a feature forms a directed ray, an axis-like pattern, multiple context regimes, a low-dimensional effect span, or a diffuse cloud.

\subsection{Logit-Induced Effect Geometry}
\label{sec:logit_induced_effect}

We formalize the geometry used throughout the FEGA diagnostics. Each removal effect is decomposed into a magnitude and a direction; directions are compared through a shared kernel; and the corresponding logit-space quantities are computed from pre-logit effects using the unembedding Gram matrix.

\indent\textbf{\textit{Magnitude-Direction Split.}}
All FEGA diagnostics are defined on the downstream logit-effect cloud. Let $\mathcal E_{j,\mathrm{logit}}=\{\Delta_{j,\mathrm{logit}}^{(c_i)}\}_{i=1}^{n_j}$ again denote the retained logit-removal effects for feature $j$, where $c_1,\ldots,c_{n_j}$ enumerate the valid contexts. For each retained effect, define its logit magnitude and normalized direction as:
$$
m_j^{(i)}=
\|\Delta_{j,\mathrm{logit}}^{(c_i)}\|_2,
\qquad
v_j^{(i)}=
\frac{\Delta_{j,\mathrm{logit}}^{(c_i)}}{m_j^{(i)}},
\qquad
\|v_j^{(i)}\|_2=1.
$$

The normalized directions $v_j^{(i)}$ capture the shape of the effect cloud after removing variation in effect magnitude. The magnitudes $m_j^{(i)}$ are used later as evidence-strength and magnitude-heterogeneity information.

\indent\textbf{\textit{Shared Directional Kernel.}}
FEGA summarizes directional agreement with a context-by-context kernel. Each entry in this kernel asks how similar two retained normalized effects are under the induced logit geometry. Let $V_j\in\mathbb R^{n_j\times |\mathcal V|}$ be the matrix whose $i$th row is $v_j^{(i)}$. The pairwise directional similarities are:

$$
(K_j)_{ik}=
\langle v_j^{(i)},v_j^{(k)}\rangle_2,
\qquad
K_j=V_jV_j^\top\in\mathbb R^{n_j\times n_j}.
$$

Positive entries indicate that two removal effects point in similar logit directions, negative entries indicate opposing directions, and entries near zero indicate little directional agreement. Thus, $K_j$ is the shared object used by the directional diagnostics below.

The non-zero eigenvalues of $K_j$ are the squared singular values of $V_j$, giving the standard dual-PCA view of the normalized logit-effect cloud \citep{DBLP:journals/neco/ScholkopfSM98}. The axis, span, residual, and effective-rank diagnostics all reuse this kernel or its centered version. Appendix~\ref{app:dual_span_spectrum} gives the full derivation.

\indent\textbf{\textit{Pre-Logit Gram Computation.}}
Since $\Delta_{j,\mathrm{logit}}^{(c)}=W_U\Delta_{j,\mathrm{pre}}^{(c)},$ the same logit inner products can be computed by pre-logit inner products without explicitly materializing expensive vocabulary-dimensional logit vectors. Define the unembedding Gram matrix:

$$
G= W_U^\top W_U\in\mathbb R^{d_{\mathrm{model}}\times d_{\mathrm{model}}}.
$$

For any two valid contexts $c_i$ and $c_k$,

$$
\langle
\Delta_{j,\mathrm{logit}}^{(c_i)},
\Delta_{j,\mathrm{logit}}^{(c_k)}
\rangle_2=
(\Delta_{j,\mathrm{pre}}^{(c_i)})^\top
G
\Delta_{j,\mathrm{pre}}^{(c_k)}.
$$

Therefore, the same magnitudes and kernel entries are computed directly using pre-logit effects as:
$$
m_j^{(i)}=\sqrt{
(\Delta_{j,\mathrm{pre}}^{(c_i)})^\top
G
\Delta_{j,\mathrm{pre}}^{(c_i)}
}, 
\qquad
(K_j)_{ik}=\frac{
(\Delta_{j,\mathrm{pre}}^{(c_i)})^\top
G
\Delta_{j,\mathrm{pre}}^{(c_k)}
}
{m_j^{(i)}m_j^{(k)}}.
$$

This is only a computational shortcut: the geometry remains the Euclidean geometry of the corresponding logit effects. Appendix~\ref{app:gram_validation} gives the validation checks and failure cases.

\subsection{Geometry Diagnostics}

\begin{figure}[t]
    \centering
    \includegraphics[width=1\linewidth]{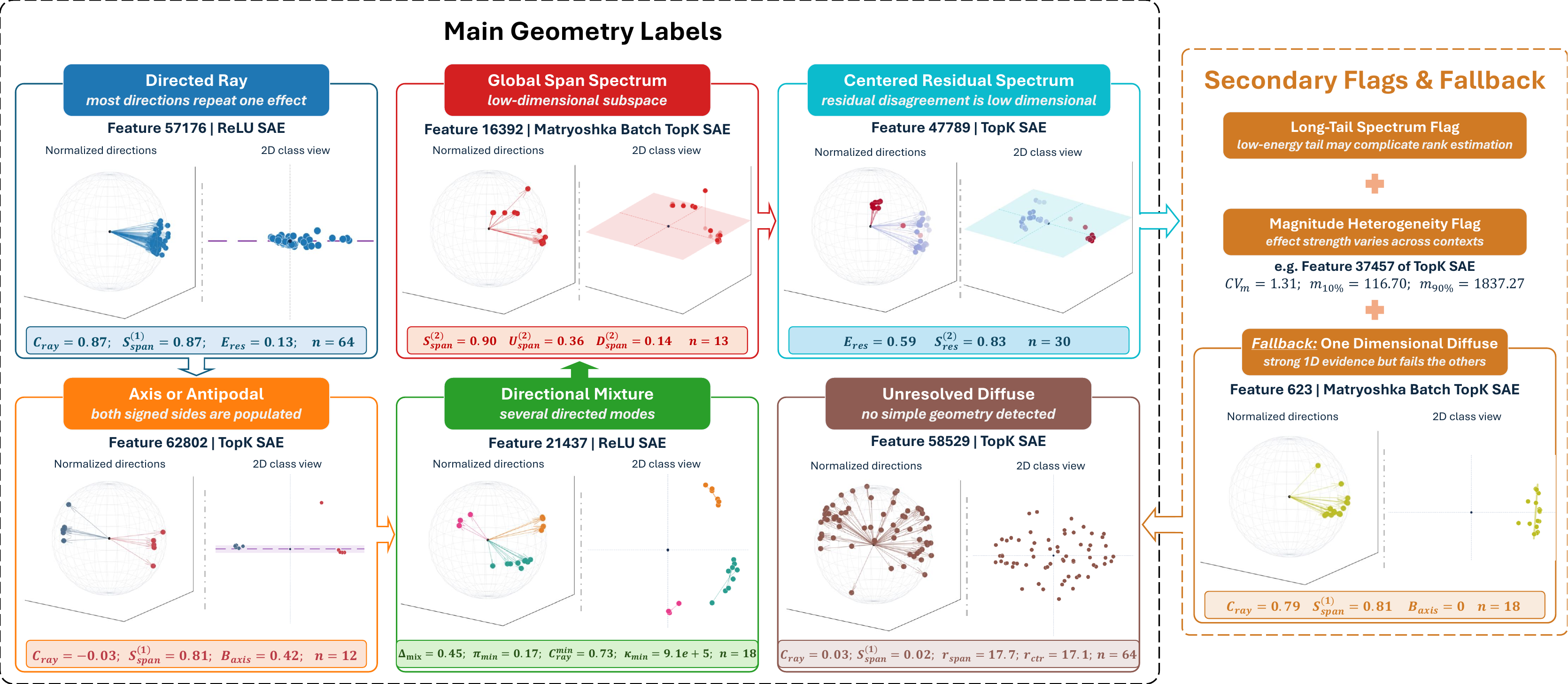}
    \caption{
    \textbf{Overview of FEGA geometry labels and secondary flags.}
    Each card shows a representative downstream logit-effect cloud after removing one SAE feature. The left panel visualizes normalized effect directions; the right panel gives a two-dimensional view for visualization only. The six main cards illustrate the strict geometry families and the unresolved outcome, rather than the complete reporting vocabulary. The orange panel shows two secondary qualifications and the one-dimensional diffuse fallback. Partial span and residual fallbacks, together with the three non-geometric terminal outcomes, are defined in Section~\ref{subsec:geometry_labels} and Appendix~\ref{app:geometry_classifier_gates}. The printed statistics are the diagnostics used by the tests.
    }
    \label{fig:geometry_labels}
\end{figure}

Starting from the normalized effect directions, FEGA tests progressively different geometric explanations: one signed direction, one unsigned axis, several directional modes, a shared low-dimensional span, and low-dimensional residual variation around the mean.

\indent\textbf{\textit{Directed-Ray Concentration.}}
The first directional question is whether the normalized effects behave like repeated observations of the same directed removal movement. Here, agreement means pointing in the same direction: opposing effects should lower the score rather than be treated as the same pattern. FEGA summarizes this alignment using the mean off-diagonal similarity in the directional kernel:

$$
C_{\mathrm{ray}}(j)=
\frac{1}{n_j(n_j-1)}
\sum_{i\neq k}
(K_j)_{ik}.
$$

High $C_{\mathrm{ray}}$ means that most retained removal effects point along a common logit direction, supporting the directed-ray family shown in Figure~\ref{fig:geometry_labels}.

Equivalently, let $S_j=\sum_i v_j^{(i)}.$ Since each $v_j^{(i)}$ is unit-normalized in logit space,

$$
C_{\mathrm{ray}}(j)=
\frac{\|S_j\|_2^2-n_j}{n_j(n_j-1)}.
$$

Appendix~\ref{app:cosine_fers_identity} gives the trace-corrected Gram form and derives this identity. Low $C_{\mathrm{ray}}$ is ambiguous: it may arise from antipodal, multimodal, low-dimensional, or diffuse structure. Negative values indicate substantial cancellation or opposition. Thus, directed-ray concentration tests one geometric hypothesis: \emph{do these removals repeatedly point in the same direction?}

\indent\textbf{\textit{Axis or Antipodal Structure.}}
A low directed-ray score may occur when effects share one axis but use opposite orientations. Contexts near $u$ and $-u$ cancel under signed similarity even though the cloud remains one-dimensional. To capture this structure, let the leading eigenpair of $K_j$ satisfy:

$$
K_j q_1=\lambda_1 q_1,
\qquad
\lambda_1\geq \lambda_2\geq\cdots.
$$

The leading eigenvector $q_1$ assigns weights to the retained contexts. Combining these weights with the sampled effect directions recovers the leading axis in logit-effect space:

$$
u_1=\frac{V_j^\top q_1}{\sqrt{\lambda_1+\epsilon}}.
$$

The corresponding signed projection scores are:

$$
s=\frac{K_jq_1}{\sqrt{\lambda_1+\epsilon}},
$$

with $s_i=\langle v_j^{(i)},u_1\rangle_2$ up to the numerical safeguard. This is the dual-PCA recovery step: diagonalize the context kernel, then map the leading component back to a logit-effect axis. Appendix~\ref{app:dual_span_spectrum} gives the full derivation and the $\lambda_1>0$ condition.

The sign of $u_1$ is arbitrary, so only the occupancy of both orientations matters. We measure it using the axis-split fraction:

$$
B_{\mathrm{axis},j}=
\frac{\min\{|\{i:s_i>0\}|,|\{i:s_i<0\}|\}}{n_j}.
$$

Axis evidence requires more than a non-zero split: it combines strong one-dimensional structure, weak directed-ray evidence, and non-negligible occupancy on both sides of the axis. This supports the axis or antipodal family shown in Figure~\ref{fig:geometry_labels}.

\indent\textbf{\textit{Directional Mixture Structure.}}
Some clouds contain several coherent directions rather than one ray or one sign-split axis. An axis contains directions near $u$ and $-u$ on the same line, whereas a directional mixture contains prototype directions that need not be collinear or antipodal.

The mixture diagnostic models the normalized logit directions $v_j^{(i)}$ with a von Mises-Fisher mixture, a distribution for unit vectors on a Euclidean sphere \citep{JMLR:v6:banerjee05a}. Because the likelihood is defined on the logit sphere, this diagnostic uses the normalized logit directions themselves (as defined in Section \ref{sec:logit_induced_effect}). For $M$ directional modes, the mixture density is:
$$
p(v_j^{(i)})=
\sum_{r=1}^M
\pi_r C_{|\mathcal V|}(\kappa_r)\exp\left(\kappa_r\mu_r^\top v_j^{(i)}\right),
$$

where $C_d(\kappa_r)$ is the vMF normalizing constant, $\mu_r$ is the prototype direction for mode $r$, $\kappa_r$ measures how tightly contexts concentrate around that prototype, and $\pi_r$ is the fitted probability mass of mode $r$, with $\pi_r\geq0$ and $\sum_{r=1}^{M}\pi_r=1$, describing how much of the effect cloud is associated with that directional regime. A context may support several modes to different degrees and is assigned to its best-matching mode only for mode-specific summaries.
\citep{JMLR:v6:banerjee05a}.

Candidate mode counts are selected using the Bayesian information criterion (BIC) \citep{schwarz1978estimating}. Bayesian information criterion (BIC) makes the proposed multi-mode explanation pay for its extra prototype directions, concentration parameters, and mixture weights. Accepted mixtures must contain more than one mode. Each mode must satisfy the fitted-mass requirement $\pi_r\geq0.10$, and each within-mode concentration must be defined from at least two hard-assigned contexts. The modes must additionally exhibit sufficient within-mode coherence and improve coherence relative to the pooled cloud. The improvement is summarized by:

$$
\Delta_{\mathrm{mix}}=\sum_{r=1}^M \pi_r C_{\mathrm{ray},r}-C_{\mathrm{ray},\mathrm{global}}.
$$

Here, $\pi_r$ describes how much probability mass the fitted mixture assigns to mode $r$, while $C_{\mathrm{ray},r}$ measures the directional coherence of the contexts whose best-matching mode is $r$; $C_{\mathrm{ray},\mathrm{global}}$ is the corresponding concentration of the full cloud. A positive $\Delta_{\mathrm{mix}}$ indicates that the proposed modes are more coherent than the full cloud. The remaining gates require valid concentration estimates and stable assignments under resampling. Appendix~\ref{app:vmf_model_selection} gives the complete fitting and model-selection protocol, the evaluation of the high-dimensional normalizer, and the assignment-stability calculation.

An accepted mixture supports multiple directional regimes in the sampled logit-effect cloud, but does not by itself establish behavioral polysemanticity. The modes may instead reflect task regimes, prompt templates, token positions, or other context variables.

\indent\textbf{\textit{Global Span Spectrum.}}
If the cloud is not explained by a ray, axis, or directional mixture, the global span tests whether all normalized effects lie within a low-dimensional logit-effect subspace. Unlike a mixture, which separates contexts into local directions, the global span summarizes the full cloud with one shared subspace.

Because $K_j$ is built from $n_j$ retained contexts, it contains at most $n_j$ non-zero eigenvalues. We treat any remaining components within the $d_{\mathrm{model}}$-dimensional readout space as having zero energy. For $\lambda_1 \geq \lambda_2 \geq \cdots$, the $k$-span sufficiency score is:

$$
S_{\mathrm{span}}^{(k)}=\frac
{\sum_{\ell=1}^k\lambda_\ell}
{\sum_\ell\lambda_\ell+\epsilon}.
$$

This score measures the fraction of directional energy explained by the first $k$ components. Since high cumulative energy alone can hide a dominant first component or a long spectral tail, FEGA also reports the component-use share:

$$ 
U_{\mathrm{span}}^{(k)}=\frac
{\lambda_k}
{\sum_\ell\lambda_\ell+\epsilon},
$$

and the post-$k$ spectral-drop ratio:

$$
D_{\mathrm{span}}^{(k)}=\frac
{\lambda_{k+1}}
{\lambda_k+\epsilon}.
$$

Here $S_{\mathrm{span}}^{(k)}$ measures whether $k$ dimensions are sufficient, $U_{\mathrm{span}}^{(k)}$ measures whether the $k$th component is meaningfully used, and $D_{\mathrm{span}}^{(k)}$ measures whether there is a clean drop after $k$. These statistics are interpreted together with effective rank (Section \ref{sec:sec_flags}). Passing the gates with $k=2$ yields a global two-dimensional directional-subspace label (Figure~\ref{fig:geometry_labels}), while passing them with $k\in\{3,4,8\}$ yields a global $k$-dimensional label. Appendix~\ref{app:illustrative_geometry_profiles} gives representative spectra.

\indent\textbf{\textit{Centered Residual Spectrum.}}
The preceding diagnostics analyze the whole cloud of normalized directions. This can miss a feature with a reliable average effect plus a smaller, systematic context-dependent deviation: the mean direction may dominate the global spectrum, while the variation around it still has low-dimensional structure. The centered residual spectrum isolates this variation by subtracting the average normalized direction before computing a spectrum. Let

$$
H=I-\frac{1}{n_j}\mathbf 1\mathbf 1^\top\in\mathbb R^{n_j\times n_j},
\qquad
K_{\mathrm{ctr}}=HK_jH.
$$

The matrix $H$ performs row-centering in context space: it subtracts the sample mean direction from each retained context. Since $K_j=V_jV_j^\top$,

$$
K_{\mathrm{ctr}}=HV_jV_j^\top H=(HV_j)(HV_j)^\top
$$

is the Gram matrix of the mean-centered logit directions. The fraction of directional energy remaining after centering is:

$$
E_{\mathrm{res}}=
\frac{\operatorname{tr}(K_{\mathrm{ctr}})}
{\operatorname{tr}(K_j)+\epsilon}.
$$

Low $E_{\mathrm{res}}$ indicates that the mean direction already explains most of the cloud. When residual energy is nontrivial, the centered eigenvalues $\eta_1\geq\eta_2\geq\cdots$ define:

$$
S_{\mathrm{res}}^{(k)}=
\frac{\sum_{\ell=1}^k\eta_\ell}
{\sum_\ell\eta_\ell+\epsilon}.
$$

The residual low-dimensional label should be driven by nontrivial residual energy together with high small-$k$ sufficiency. FEGA makes this claim for $k \in \{2,3,4\}$; a one-dimensional residual may be noted as a descriptive fallback, but is not treated as a supported residual-dimensionality claim. When one of these dimensions is supported, the cloud has an average direction and its context-dependent deviations occupy a small subspace. Tiny residual energy should not drive a residual-structure claim. Appendix~\ref{app:centered_residual_spectrum} gives the derivation.

\begin{table}[t]
\centering
\footnotesize
\setlength{\tabcolsep}{3.5pt}
\renewcommand{\arraystretch}{1}
\begin{tabular}{@{}P{0.17\linewidth}P{0.22\linewidth}P{0.28\linewidth}P{0.23\linewidth}@{}}
\arrayrulecolor{black}
\hline
\textbf{Diagnostic} & \textbf{What It Detects} & \textbf{Key Readout} & \textbf{Guardrail} \\
\hline
\arrayrulecolor{gray!25}
Directed ray & One shared direction & $C_{\mathrm{ray}}$ & Low score has many causes. \\
\hline
Axis split & One axis, two signs & $B_{\mathrm{axis},j}$ & Sign split is not temporal flipping. \\
\hline
vMF mixture & Several coherent directions & Accepted modes $M$, coherence gain $\Delta_{\mathrm{mix}}$ & Modes must be populated and stable. \\
\hline
Span spectrum & One shared low-D subspace & $S_{\mathrm{span}}^{(k)}$, $U_{\mathrm{span}}^{(k)}$, $D_{\mathrm{span}}^{(k)}$ & High energy alone can hide tails. \\
\hline
Residual spectrum & Low-D variation around the mean & $E_{\mathrm{res}}$, $S_{\mathrm{res}}^{(k)}$ & Residual energy must be nontrivial. \\
\hline
Secondary flags & Strength and failure modes & $\mathrm{CV}_{m,j}$, long-tail, instability flags & Flags qualify the label. \\
\hline
Selected-family stability & Confidence in the selected result & Family-specific intervals, assignment agreement, leave-out, sample-size, or angle evidence & Availability is reported separately. \\
\hline
\arrayrulecolor{black}
Reporting label & Final summary family & Primary label plus flags & Descriptive, not mechanistic. \\
\hline
\arrayrulecolor{black}
\end{tabular}
\caption{\textbf{FEGA diagnostic map.} Each row tests one property of the same effect cloud. The key readout gives the statistic used for the test; the guardrail states what the statistic alone should not be taken to imply.}
\label{tab:fega_diagnostic_map}
\end{table}

\subsection{Evidence Strength and Secondary Flags}
\label{sec:sec_flags}

The diagnostics above identify the geometric family best supported by the full cloud. FEGA then evaluates only the stability evidence relevant to that selected family and, where applicable, its selected dimension. Sample size, spectral tails, and magnitude heterogeneity remain companion summaries; none of these checks replaces the full-sample label.

\indent\textbf{\textit{Effective Rank.}}
The span statistics evaluate a chosen value of $k$. Effective rank provides a complementary summary of how many components the relevant spectrum uses overall. For the global span diagnostic, this spectrum is given by the eigenvalues of $K_j$; for the centered residual diagnostic, it is given by the eigenvalues of $K_{\mathrm{ctr}}$. Thus, effective rank acts as a safeguard: it helps distinguish a clean low-dimensional spectrum from a spectrum with one large component plus a long tail.

For a non-negative spectrum $s_1,\ldots,s_r$, define:

$$
p_\ell=
\frac{s_\ell}{\sum_{q=1}^r s_q+\epsilon}.
$$

The entropy and participation-ratio ranks are:

$$
r_{\mathrm{ent}}=
\exp\left(-\sum_\ell p_\ell\log(p_\ell+\epsilon)\right),
\qquad
r_{\mathrm{PR}}=
\frac{1}{\sum_\ell p_\ell^2}.
$$

Ignoring the numerical safeguard, both equal $k$ for a uniform $k$-component spectrum. Participation-ratio rank emphasizes components carrying substantial energy, while entropy rank is more sensitive to long low-energy tails. FEGA uses $r_{\mathrm{PR}}$ as the more conservative dimensionality summary and $r_{\mathrm{ent}}$ as a long-tail warning. Neither statistic defines a geometry family on its own; instead, they qualify span and residual claims made from $S_{\mathrm{span}}^{(k)}$ and $S_{\mathrm{res}}^{(k)}$.

\indent\textbf{\textit{Magnitude Heterogeneity.}}
The primary geometry family describes the normalized removal directions $v_j^{(i)}$. Because normalization removes effect size, FEGA reports magnitude heterogeneity separately using the logit-effect magnitudes $m_j^{(i)}$ defined in Section~\ref{sec:fega_setup}:

$$
\mathrm{CV}_{m,j}=
\frac{\operatorname{std}_i(m_j^{(i)})}
{\operatorname{mean}_i(m_j^{(i)})}.
$$

High $\mathrm{CV}_{m,j}$ adds a magnitude-instability flag, indicating that effect strength varies across contexts even when the normalized directions support a coherent geometry. This flag qualifies the primary label but does not replace it. For example, a feature can be assigned a directed-ray or low-dimensional-span label while still having unstable effect magnitudes. Long-tail spectral flags similarly qualify low-dimensional summaries. Sensitivity to omitted contexts is incorporated into the selected-family stability result described below. Appendix~\ref{app:illustrative_geometry_profiles} gives examples, and Appendix~\ref{app:geometry_classifier_gates} lists the flag triggers.

\indent\textbf{\textit{Stability and Evidence Strength.}}
FEGA reports insufficient effect evidence when fewer than eight valid non-zero contexts remain. Otherwise, it first selects the most specific family supported by the full cloud and, for a dimensioned family, the smallest supported dimension. Stability analysis then considers only evidence relevant to that selected result.

The relevant evidence may be a ray-concentration interval, family-local leave-out and sample-size checks, a principal angle at the selected dimension, or the assignment agreement already computed for an accepted directional mixture. Stability changes the confidence attached to the result, not its family or dimension. Observed instability and unavailable evidence are reported separately, while fallback and terminal outcomes are marked as not evaluated. Appendix~\ref{app:fega_stability_protocols} specifies the family-specific procedures, and Appendix~\ref{app:geometry_classifier_gates} lists the reporting thresholds. The thresholds are reporting gates rather than universal constants.

\begin{figure}[t]
    \centering
    \includegraphics[width=1\linewidth]{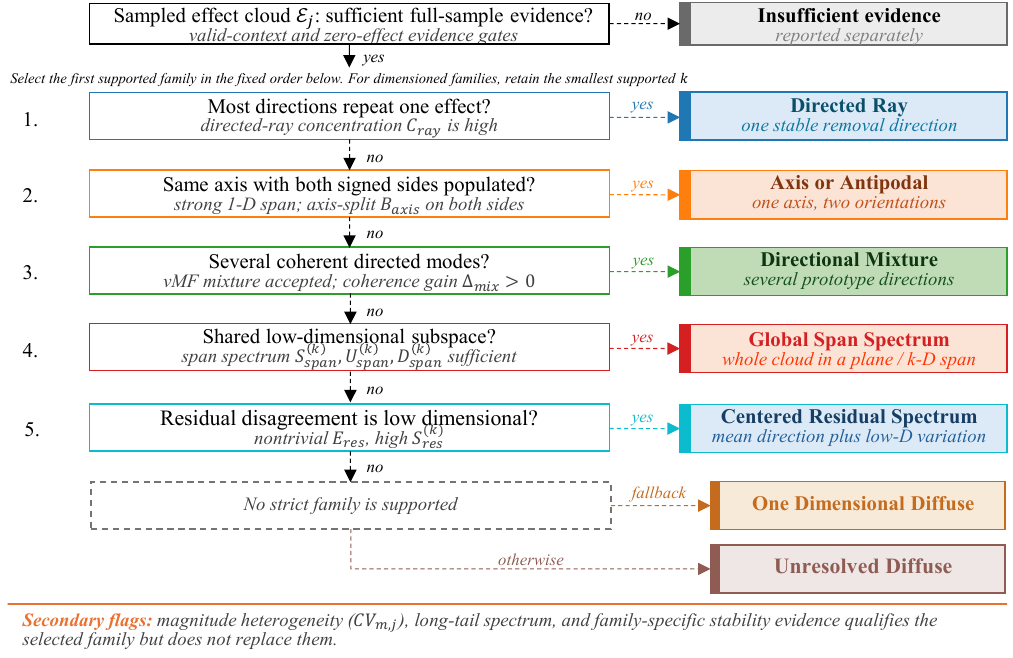}
    \caption{\textbf{FEGA label assignment and qualification.} Full-sample diagnostics select the first supported geometry family and, where applicable, its smallest supported dimension. Family-specific stability tests then qualify this selection; directional-mixture acceptance already includes assignment stability.}
    \label{fig:geometry_algo}
\end{figure}

\subsection{Geometry Labels}
\label{subsec:geometry_labels}

Figure~\ref{fig:geometry_algo} summarizes the FEGA label-assignment procedure. FEGA records the full point-diagnostic profile for each feature together with stability evidence for the selected family, but assigns one primary geometry label to that feature's sampled logit-effect cloud for empirical summaries.

Clouds with insufficient valid effects are excluded before geometry selection. For each remaining cloud, FEGA considers the families in the fixed order directed ray, axis or antipodal structure, accepted directional mixture, global low-dimensional span, and centered residual low-dimensional structure, as shown in Figure~\ref{fig:geometry_algo}. Ray, axis, span, and residual selection use full-sample diagnostics; directional-mixture acceptance is the sole exception because it already includes its standalone assignment-stability audit. The first supported family is retained, and dimensioned families use the smallest supported $k$. Subsequent family-specific stability checks qualify this fixed choice rather than searching for a replacement. This priority gives more specific explanations precedence: a directed ray, for example, is also low-dimensional, but is more informative than a generic span. Appendix~\ref{app:geometry_classifier_gates} gives the exact gates and priority rules.

If no strict family passes, FEGA reports the first descriptive fallback supported by the full cloud: one-dimensional diffuse evidence, the first anchored global span, the first anchored centered residual span, or unresolved high-dimensional or diffuse evidence. Directed rays, axes, and accepted directional mixtures are strict selections rather than separate fallback branches. Fallback outcomes describe the nearest supported structure and are not subjected to the selected-family stability protocol.

The three terminal outcomes answer different questions. \emph{Insufficient effect evidence} means that the initial evidence gate failed; \emph{geometry metrics unavailable} means that the evidence gate passed but every family test lacked the quantities needed for evaluation; and \emph{undefined geometry} means that family tests were attempted but neither a strict family nor a descriptive fallback was supported.

%% file: sections/analysis.tex
\begin{figure}[t]
    \centering
     \includegraphics[width=1\linewidth]{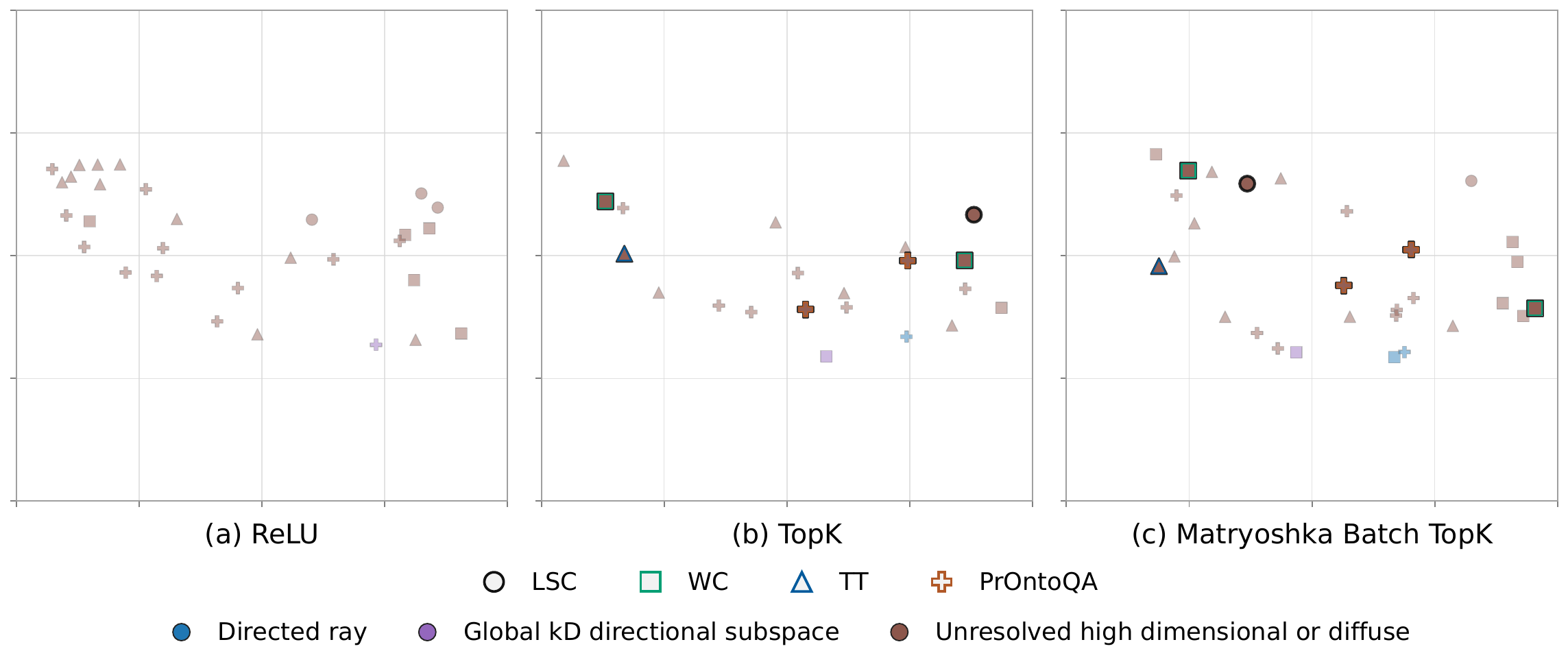}
    \caption{\textbf{FEGA atlas for pointer-like features.} UMAP projections of downstream logit-effect geometry across four in-context tasks. Marker shape denotes the task and color denotes the FEGA label; features with undefined geometry are omitted. Features shared across all four tasks are shown at higher opacity.}
    \label{fig:algorithmic-results}
\end{figure}

\section{Geometric Analysis of Downstream Effects}
\label{sec:geometric_analysis}

Having formalized FEGA, we apply it to isolated features across our evaluation spectrum, from synthetic pattern matching to factual attribute retrieval. This lets us systematically ask how downstream effect geometry varies with a feature's computational role.

\begin{table*}[t]
    \centering
    \footnotesize
    \setlength{\tabcolsep}{4pt}
    \begin{tabular}{lcccccccccc}
        \toprule
        \textbf{Task} & \textbf{Selected} & \textbf{Undef.} & \textbf{Unres.} & \textbf{G-$k$D} & \textbf{Res.-lowD} & \textbf{1D diff.} & \textbf{Ray} & \textbf{G-2D} & \textbf{Axis} & \textbf{Multi} \\
        \arrayrulecolor{black}
        \hline
        \rowcolor{gray!15} \multicolumn{11}{c}{\textbf{ReLU SAE}} \\ \hline
        LSC & 5 & 2 & 3 & 0 & 0 & 0 & 0 & 0 & 0 & 0 \\
        \arrayrulecolor{gray!25}
        \hline
        WC & 25 & 20 & 5 & 0 & 0 & 0 & 0 & 0 & 0 & 0 \\
        \arrayrulecolor{gray!25}
        \hline
        PrOntoQA & 78 & 66 & 11 & 1 & 0 & 0 & 0 & 0 & 0 & 0 \\
        \arrayrulecolor{gray!25}
        \hline
        TT & 27 & 17 & 10 & 0 & 0 & 0 & 0 & 0 & 0 & 0 \\
        \arrayrulecolor{black}
        \hline
        \rowcolor{gray!15} \multicolumn{11}{c}{\textbf{TopK SAE}} \\ \hline
        LSC & 5 & 4 & 1 & 0 & 0 & 0 & 0 & 0 & 0 & 0 \\
        \arrayrulecolor{gray!25}
        \hline
        WC & 9 & 5 & 3 & 1 & 0 & 0 & 0 & 0 & 0 & 0 \\
        \arrayrulecolor{gray!25}
        \hline
        PrOntoQA & 22 & 13 & 8 & 0 & 0 & 0 & 1 & 0 & 0 & 0 \\
        \arrayrulecolor{gray!25}
        \hline
        TT & 14 & 7 & 7 & 0 & 0 & 0 & 0 & 0 & 0 & 0 \\
        \arrayrulecolor{black}
        \hline
        \rowcolor{gray!15} \multicolumn{11}{c}{\textbf{Matryoshka Batch TopK SAE}} \\ \hline
        LSC & 4 & 2 & 2 & 0 & 0 & 0 & 0 & 0 & 0 & 0 \\
        \arrayrulecolor{gray!25}
        \hline
        WC & 14 & 5 & 7 & 1 & 0 & 0 & 1 & 0 & 0 & 0 \\
        \arrayrulecolor{gray!25}
        \hline
        PrOntoQA & 20 & 10 & 9 & 0 & 0 & 0 & 1 & 0 & 0 & 0 \\
        \arrayrulecolor{gray!25}
        \hline
        TT & 16 & 8 & 8 & 0 & 0 & 0 & 0 & 0 & 0 & 0 \\
        \arrayrulecolor{black}
        \bottomrule
    \end{tabular}
    \caption{\textbf{Primary FEGA labels for pointer-like features.} Geometry labels for features isolated from the four ICL tasks across Gemma-2-2B SAE variants of width $65$k. \textit{Selected} is the number of features active across at least $90\%$ examples and on at least $90\%$ queries in at least $90\%$ prompt families per ICL task; \textit{Undef.} denotes eligible features for which family tests were attempted, but neither a strict family nor a descriptive fallback was supported. Other columns denote unresolved/diffuse (\textit{Unres.}), global $k$-dimensional subspace (\textit{G-$k$D}), residual low-dimensional (\textit{Res.-lowD}), one-dimensional diffuse (\textit{1D diff.}), directed ray (\textit{Ray}), global 2D subspace (\textit{G-2D}), axis or antipodal (\textit{Axis}), and directional mixture (\textit{Multi}).}
    \label{tab:algorithmic-geometry-counts}
\end{table*}

\subsection{The Geometry of Pointer-Like Features}
\label{subsec:geometry_pointer}

We first examine features isolated from the four ICL tasks: LSC, WC, PrOntoQA, and TT. We call these features pointer-like candidates as they recur across model-correct examples in tasks involving prompt-local copying, lookup, rule completion, or schema following.

Figure~\ref{fig:algorithmic-results} and Table~\ref{tab:algorithmic-geometry-counts} show a clear pattern. Of the $239$ selected features, $159$ are classified as having undefined geometry. Among the $80$ mapped cases, $74$ are classified as unresolved high-dimensional or diffuse. Only three form directed rays, and three occupy global low-dimensional subspaces. Thus, when pointer-like candidates have sufficient evidence for geometric analysis, stable low-dimensional structures are rare.

This pattern is consistent with their context-dependent targets. A feature may support a similar operation across prompts, but copying ``apple'' and copying ``car'', for instance, require effects on different output logits. As the relevant value changes, the downstream effect can change with it. A shared functional role therefore need not produce a shared direction in logit space.

TT provides a similar example. The prompt specifies the target language, but the correct translation changes with the source word. A feature shared across TT examples therefore contributes to predicting many different output tokens, so its effect need not follow a fixed direction in logit space.

\begin{takeaway}
\textbf{Mapped pointer-like effects are overwhelmingly diffuse.}
Among pointer-like candidates with a mapped geometry, stable low-dimensional structures are rare, consistent with their effects changing as the prompt-local target changes.
\end{takeaway}

\subsection{The Geometry of Value-Like Features}
\label{subsec:geometry_value}

We next evaluate features isolated from the RAVEL city-country attribute task. Since these latents are selected by differential binary masking to support attribute editing (e.g., changing a city's country attribute), they are natural candidates for stable conceptual control vectors. FEGA tests whether such value-like features produce a consistent one-dimensional downstream effect across contexts.

\begin{figure}[tbp]
    \centering
    \includegraphics[width=\linewidth]{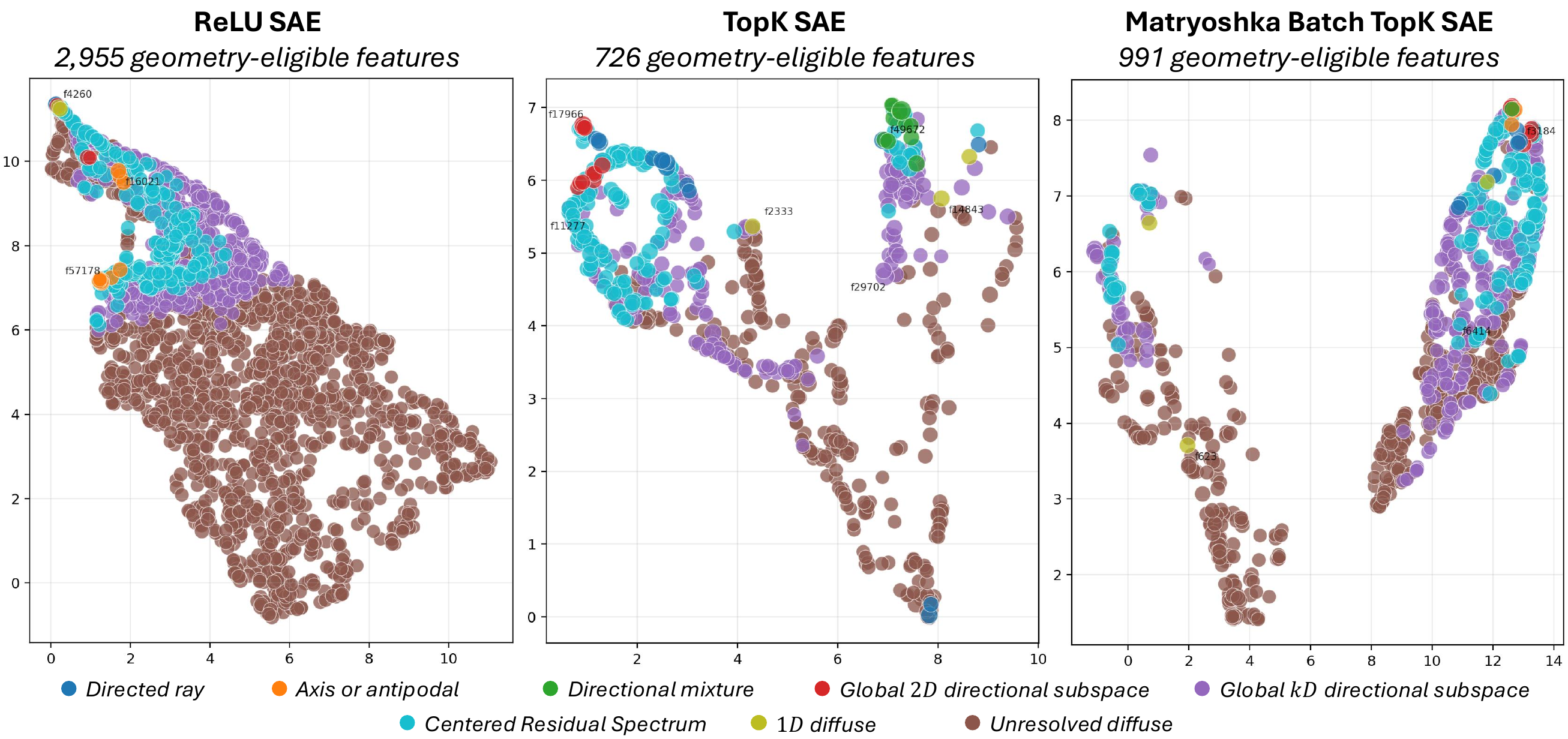}
    \caption{\textbf{FEGA atlas for value-like features.} UMAP projections of the RAVEL features for which FEGA assigns a primary geometry label. Features with insufficient effect evidence or undefined geometry are omitted from the atlas and retained in the complete counts in Table~\ref{tab:ravel-geometry-counts}. Colors indicate the primary FEGA label.}
    \label{fig:ravel-results}
\end{figure}

Figure~\ref{fig:ravel-results} shows the diversity of the mapped RAVEL geometries, but it does not display the entire selected population. Features with insufficient effect evidence and features for which no geometry family is supported are omitted from the atlas and reported separately in Table~\ref{tab:ravel-geometry-counts}. Among the remaining cases, isolating a factual attribute at the representation level still does not imply a single downstream ray.

Table~\ref{tab:ravel-geometry-counts} quantifies this pattern. Of the 12{,}900 ReLU, 2{,}828 TopK, and 3{,}654 Matryoshka Batch TopK features selected by differential binary masking, 5{,}185, 1{,}661, and 1{,}904, respectively, lack sufficient effect evidence for a geometry claim. Among the geometry-eligible features, undefined geometry accounts for 4,760 of 7,715 ReLU features, 441 of 1,167 TopK features, and 759 of 1,750 Matryoshka Batch TopK features. Unresolved high-dimensional or diffuse evidence accounts for a further 1,860, 277, and 452 features, respectively. 
Together, the G-$k$D, residual-low-dimensional, 1D-diffuse, and G-2D categories contain 1{,}078 ReLU features, 424 TopK features, and 530 Matryoshka Batch TopK features, corresponding to 14.0\%, 36.3\%, and 30.3\% of the eligible populations. Directed rays remain much rarer, with 5, 14, and 6 features, respectively.


This geometric heterogeneity is consistent with the complexity of natural language generation. Even if an individual SAE feature encapsulates a concept such as the country ``France'', ablating it from the model's residual stream may alter vocabulary probabilities along multiple axes. Depending on the grammar and phrasing of the prompt, suppressing that feature might boost probabilities for competing geographic entities (e.g., ``Japan''), alter regional language associations (e.g., shifting toward ``Spanish''), or produce a localized syntactic adjustment. In structured cases, the feature's downstream effects therefore occupy a low-dimensional subspace rather than a single direction. This shows that understanding \textit{what} conceptual knowledge a feature encodes is distinct from predicting \textit{how} its intervention will geometrically perturb the output space.

\begin{takeaway}
\textbf{Value-like candidates exhibit low-dimensional structure more often than pointer-like candidates.}
Their structured effects usually span multiple directions, while directed rays remain rare.
\end{takeaway}

\begin{table}[t]
  \centering
  \footnotesize
  \setlength{\tabcolsep}{3.5pt}
  \begin{tabular}{lrrrrrrrrrrr}
      \toprule
      \textbf{SAE}
      & \textbf{Selected}
      & \textbf{Insuff.}
      & \textbf{Undef.}
      & \textbf{Unres.}
      & \textbf{G-$k$D}
      & \makecell[r]{\textbf{Res.-}\\\textbf{lowD}}
      & \makecell[r]{\textbf{1D-}\\\textbf{diff.}}
      & \textbf{Ray}
      & \textbf{G-2D}
      & \textbf{Axis}
      & \textbf{Multi} \\
      \arrayrulecolor{black}
      \midrule
      ReLU
      & 12{,}900
      & 5{,}185
      & 4{,}760
      & 1{,}860
      & 795
      & 278
      & 2
      & 5
      & 3
      & 12
      & 0 \\
      \arrayrulecolor{gray!25}
        \hline
      TopK
      & 2{,}828
      & 1{,}661
      & 441
      & 277
      & 243
      & 165
      & 3
      & 14
      & 13
      & 0
      & 11 \\
      \arrayrulecolor{gray!25}
        \hline
      \makecell[l]{Matryoshka\\Batch TopK}
      & 3{,}654
      & 1{,}904
      & 759
      & 452
      & 352
      & 164
      & 3
      & 6
      & 11
      & 2
      & 1 \\
      \arrayrulecolor{black}
      \bottomrule
  \end{tabular}
  \caption{\textbf{Primary FEGA labels for value-like features.}  Geometry outcomes for features selected from the RAVEL city-country task. \textit{Selected} is the number of positive MDBM-mask entries. \textit{Insuff.} denotes features with fewer than eight valid nonzero effects. \textit{Undef.} denotes eligible features for which family tests were attempted, but neither a strict family nor a descriptive fallback was supported. The remaining columns denote unresolved high-dimensional or diffuse geometry (\textit{Unres.}), global $k$-dimensional subspaces (\textit{G-$k$D}), residual low-dimensional structure (\textit{Res.-lowD}), one-dimensional diffuse structure (\textit{1D diff.}), directed rays (\textit{Ray}), global two-dimensional subspaces (\textit{G-2D}), axes or antipodal structure (\textit{Axis}),
  and directional mixtures (\textit{Multi}).}
  \label{tab:ravel-geometry-counts}
\end{table}

%% file: sections/conclusion.tex
\section{Discussion and Conclusion}
\label{sec:discussion_conclusion}

A common assumption in SAE-based steering is that an isolated feature provides a stable, one-dimensional control direction. Our results show that such directions are rare. Value-like features tied to factual attributes exhibit low-dimensional structure more often than pointer-like features, but many cases remain undefined or unresolved, and directed rays are uncommon. Thus, even when a factual intervention has a structured effect, it can perturb multiple output directions rather than following a single steering vector. Among pointer-like candidates with mapped geometries, effects are overwhelmingly diffuse or high-dimensional, consistent with their downstream directions changing as the prompt-local target changes.

These differences have direct implications for steering. When a value-like effect spans a low-dimensional subspace, steering along a single direction may also alter related output probabilities. The prompt-local operations studied here pose a stronger challenge for static steering: their effects suggest that fixed, context-invariant vectors may be insufficient for controlling copying, binding, and similar operations. Such interventions may instead need to adapt to the prompt and to the value currently being operated on.

These findings also affect how SAEs should be trained and evaluated. Cross-task overlap is greater for TopK and Matryoshka Batch TopK than for ReLU SAEs, suggesting that SAE architecture influences how features associated with prompt-local operations are organized. However, no architecture consistently yields the largest candidate sets, and the selected sets remain architecture-dependent. Our results therefore motivate training objectives and evaluations that complement factual recall with pointer-like functions and other context-dependent computations.

Overall, SAE interpretation should move beyond asking only \textit{what} a feature represents. For reliable steering, auditing, and editing, we must also ask \textit{how} intervening on that feature geometrically perturbs the model's downstream output space.

%% file: sections/limitations.tex
\section{Limitations and Future Work}
\label{sec:limitations}

FEGA provides a framework for evaluating downstream feature stability, but our study has several boundaries.

First, all experiments use Gemma-2-2B with $65$k-width SAEs applied to the post-layer-12 residual stream. Although we compare ReLU, TopK, and Matryoshka Batch TopK variants, feature organization may change across model scales, SAE widths, or intervention layers. Larger models may also require different SAE capacities to isolate pointer-like mechanisms. Whether the predominance of diffuse geometry among mapped pointer-like candidates persists at larger scales is therefore an open empirical question.

Second, our task spectrum uses structured prompts designed to separate computational roles. LSC and RAVEL represent useful endpoints: prompt-local copying on one side and factual attribute editing on the other. In more naturalistic generation, pointer-like routing and value-like retrieval may interact within the same computation. Future work should apply FEGA to less constrained settings, such as multi-step or chain-of-thought generation, where prompt-local operations and semantic knowledge are jointly used.

Our conclusions are also conditional on the features selected by each task and on the at most $64$ active contexts retained per feature. Moreover, many selected features either lack sufficient effect evidence or do not receive a defined geometry label. Our claims about the distribution of mapped geometries should therefore not be extended to these cases. FEGA describes the geometry observed for a feature over its retained context sample; it does not claim that the same label must hold over every possible prompt. Stability analyses measure sensitivity within this sample but cannot eliminate the underlying selection boundary.

Finally, our measurements are taken at the target-position logit readout. This choice is behaviorally motivated, since logits determine the model's next-token distribution, but it also aggregates all transformations downstream of the SAE site. A feature with diffuse logit effects may still produce a low-dimensional local perturbation near that site. Applying FEGA at successive layers could reveal whether and where a localized feature effect disperses into a high-dimensional vocabulary-level effect.

%% file: sections/acknowledgement.tex
\section*{Acknowledgments}
\label{sec:ack}

P. Hoang, S. Dutta, and I. Gurevych acknowledge the support of the LOEWE Distinguished Chair ``Ubiquitous Knowledge Processing'', LOEWE initiative, Hesse, Germany (Grant Number: LOEWE/4a//519/05/00.002
(0002)/81). T. Chakraborty acknowledges the support of the Rajiv Khemani Young Faculty Chair Professorship in AI and the NVIDIA Academic Grant Program. A. Chatterjee acknowledges the support of the Google PhD Fellowship.

%% file: sections/appendix/tasks.tex

%% file: sections/appendix/metric_derivation.tex
\section{Sign Convention for Removal Effects}
\label{app:removal_contribution_sign_invariance}

FEGA orients each effect as the downstream change caused by removing a feature. For context \(c\), the logit-space removal effect is:

\[
\Delta_j^{(c)} =
\phi_\ell^{(c)}(\hat h_{\ell,j\downarrow}^{(c)}) -
\phi_\ell^{(c)}(\bar h_\ell^{(c)}),
\]

where, \(\bar h_\ell^{(c)}=D(z^{(c)})\) is the full SAE reconstruction and \(\hat h_{\ell,j\downarrow}^{(c)}=D(z^{(c)}-z_j^{(c)}e_j)\) is the reconstruction with feature \(j\) zeroed. Thus,  \(\Delta_{j,t}^{(c)}>0\) means that removing feature \(j\) raises token \(t\)'s logit, so the retained feature had been suppressing it; \(\Delta_{j,t}^{(c)}<0\) means removal lowers the logit, so the feature had been supporting it. Before unembedding, the corresponding removal effect is \(\delta_j^{(c)}=r_\ell^{(c)}(\hat h_{\ell,j\downarrow}^{(c)})-r_\ell^{(c)}(\bar h_\ell^{(c)})\), with \(\Delta_j^{(c)}=W_U\delta_j^{(c)}\).

This sign fixes the interpretation of positive and negative token changes, but not FEGA's geometry. Let \(v_i=\Delta_j^{(c_i)}/\|\Delta_j^{(c_i)}\|_2\) be the normalized logit-effect direction for feature \(j\) in context \(c_i\). If one instead used the opposite convention, describing what the retained feature contributed before removal, every direction in the feature cloud would flip: \(\tilde v_i=-v_i\). Hence \(\tilde v_i^\top \tilde v_k=(-v_i)^\top(-v_k)=v_i^\top v_k\). The same holds when computing logit geometry from pre-logit effects with \(G=W_U^\top W_U\), since \((-\delta_i)^\top G(-\delta_k)=\delta_i^\top G\delta_k\).

Thus,  magnitudes, pairwise similarities, spectra, effective-rank summaries, and FEGA geometry labels are unchanged by flipping all contexts for the same feature. The sign only matters for externally oriented quantities, such as positive versus negative token effects, agreement with a target direction, or signed intervention scores. Throughout the paper, FEGA uses the removal orientation above.

\section{Readout-Gram Equivalence and Validation}
\label{app:gram_validation}

FEGA geometry is defined in logit space, but most diagnostics require only logit-space inner products. We, therefore, retain the pre-logit removal effects in their original context order rather than storing vocabulary-dimensional logit effects. A context enters the effect cloud only when $\sqrt{\left(\delta_j^{(c)}\right)^\top G\delta_j^{(c)}}>10^{-12}.$ This set is formed once and used throughout our analysis. Since $\Delta_j^{(c)}=W_U\delta_j^{(c)}$ and $G=W_U^\top W_U$, for any retained contexts $c_i$ and $c_k$,

\[
\big\langle \Delta_j^{(c_i)}, \Delta_j^{(c_k)} \big\rangle_2 =
\big(W_U\delta_j^{(c_i)}\big)^\top
\big(W_U\delta_j^{(c_k)}\big) =
(\delta_j^{(c_i)})^\top G \delta_j^{(c_k)} .
\]

Thus,  any diagnostic depending only on logit-space inner products can be computed exactly in pre-logit coordinates through the unembedding Gram matrix, including magnitudes, cosines, pairwise kernels, directed-ray concentration, span and centered spectra, and effective-rank summaries. 

\begin{table}[t]
  \centering
  \footnotesize
  \setlength{\tabcolsep}{4pt}
  \renewcommand{\arraystretch}{0.95}
  \begin{tabular}{@{}lrrr@{}}
  \hline
  Quantity & $\mathrm{atol}$ & $\mathrm{rtol}$ & Max. absolute error \\
  \hline
  Linear-logit reconstruction & $10^{-6}$ & $10^{-6}$ & $0$ \\
  Logit norm & $10^{-6}$ & $10^{-5}$ & $7.67\times10^{-13}$ \\
  Logit inner product & $10^{-5}$ & $10^{-5}$ & $1.17\times10^{-10}$ \\
  Cosine & $10^{-5}$ & $10^{-5}$ & $8.92\times10^{-8}$ \\
  $C_{\mathrm{ray}}$ & $10^{-4}$ & $10^{-4}$ & $5.59\times10^{-9}$ \\
  $S_{\mathrm{span}}^{(1)}$ & $10^{-4}$ & $10^{-4}$ & $0$ \\
  $S_{\mathrm{res}}^{(1)}$ & $10^{-4}$ & $10^{-4}$ & $0$ \\
  \hline
  \end{tabular}
  \caption{\textbf{Gram--logit equivalence on the bounded validation sample.} Each entry is the largest
  absolute discrepancy across the eight selected features and their eight retained contexts; kernel
  comparisons include all context pairs. A comparison passes when $|a-b|\leq\mathrm{atol}+
  \mathrm{rtol}|b|$. All seven comparisons passed. Model replay used bfloat16 arithmetic, while the
  Gram and equivalence calculations used float64.}
  \label{tab:gram_logit_equivalence}
\end{table}

We evaluated this equivalence on a deterministic sample from the Gemma-2-2B ReLU SAE at layer 12 (width $2^{16}$, trainer 0) for the RAVEL city--country task. The sample contains the eight lowest-index eligible features, $1,4,17,26,44,47,50,$ and $56$, and the first eight retained contexts for each feature. As Table~\ref{tab:gram_logit_equivalence} shows, the Gram and explicit linear-logit calculations agreed for every comparison under the stated tolerances; the largest absolute difference was $8.92\times10^{-8}$, for a cosine. In contrast, the returned post-softcap logit differences departed from the corresponding linear-logit effects by as much as $1.44$, and are therefore not used to define FEGA geometry.

\section{Directed-Ray Concentration Fast Identity}
\label{app:cosine_fers_identity}

Fix a feature \(j\), write \(v_i=v_j^{(i)}\) for its \(n=n_j\) normalized logit-effect directions, and let \(K=K_j\) be the directional kernel with entries \(K_{ik}=v_i^\top v_k\). The directed-ray score is the mean ordered off-diagonal similarity,

\[
C_{\mathrm{ray}}(j) =
\frac{1}{n(n-1)}
\sum_{i\neq k} v_i^\top v_k =
\frac{\mathbf 1^\top K\mathbf 1-\operatorname{tr}(K)}
{n(n-1)}.
\]

The second equality subtracts the diagonal terms from the full kernel sum. This trace-corrected form is exact for the stored rows. When effects are stored in pre-logit form, the same entries \(K_{ik}\) are computed using the Gram metric from Appendix~\ref{app:gram_validation}.

The fast formula follows by summing the directions. Let \(S_j=\sum_{i=1}^n v_i\). Since \(\|S_j\|_2^2=\sum_{i,k}v_i^\top v_k=\mathbf 1^\top K\mathbf 1\), exact unit normalization gives \(\operatorname{tr}(K)=n\), and hence

\[
C_{\mathrm{ray}}(j)=
\frac{\|S_j\|_2^2-n}{n(n-1)}.
\]

Thus,  the fast formula is the trace-corrected identity with the diagonal replaced by \(n\). If numerical safeguards make row norms only approximately one, we use the trace-corrected form instead.

This identity also gives the intended interpretation: \(C_{\mathrm{ray}}(j)\) is high when normalized removal effects repeatedly point in the same direction. A low value only shows that a shared directed component is weak; it does not distinguish antipodal, multimodal, low-dimensional, or diffuse structure.

\section{Dual Span Spectrum}
\label{app:dual_span_spectrum}

Fix a feature \(j\), and let \(V_j\in\mathbb R^{n_j\times |\mathcal V|}\) have rows \(v_j^{(i)}\), the normalized logit-effect directions for the valid contexts. The directional kernel is \(K_j=V_jV_j^\top\), with entries \((K_j)_{ik}=(v_j^{(i)})^\top v_j^{(k)}\). When effects are stored in pre-logit form, Appendix~\ref{app:gram_validation} gives the same entries using the unembedding Gram matrix; the geometry is still the Euclidean geometry of the induced logit directions.

The spectrum of \(K_j\) is the dual-PCA spectrum of the effect cloud. If \(V_j=U\Sigma Q^\top\) is an SVD, then

\[
K_j=
V_jV_j^\top=
U\Sigma^2U^\top,
\qquad
V_j^\top V_j=
Q\Sigma^2Q^\top .
\]

Thus, the nonzero eigenvalues of $K_j$ are the squared singular values of the normalized logit-direction matrix. The $n_j \times n_j$ context kernel therefore contains the complete nonzero spectrum, allowing FEGA to diagonalize this kernel instead of constructing principal directions in vocabulary-sized logit space \citep{DBLP:journals/neco/ScholkopfSM98}. Because the effects pass through a $d_{\mathrm{model}}$-dimensional readout, spectral components not represented by $K_j$ but still within that space are assigned zero energy; indices beyond $d_{\mathrm{model}}$ do not denote available readout directions. 

For $\lambda_1 \geq \lambda_2 \geq \cdots$, the span sufficiency score:

\[
S_{\mathrm{span}}^{(k)}(j)=
\frac{\sum_{\ell=1}^k \lambda_\ell}
{\sum_\ell \lambda_\ell+\epsilon}
\]

is the fraction of logit-direction energy captured by the top \(k\) principal directions. This score asks whether \(k\) directions are enough; component-use shares and post-\(k\) spectral-drop ratios are then used to avoid mistaking a dominant first component or a long spectral tail for a clean \(k\)-dimensional cloud.

The leading axis used by the axis diagnostic is recovered from the same kernel. For an eigenpair \(K_jq_\ell=\lambda_\ell q_\ell\) with \(\lambda_\ell>0\), define:

\[
u_\ell =
\frac{V_j^\top q_\ell}{\sqrt{\lambda_\ell+\epsilon}},
\qquad
s_\ell =
V_j u_\ell =
\frac{K_jq_\ell}{\sqrt{\lambda_\ell+\epsilon}} .
\]

Here \(u_\ell\) is the recovered logit-space principal direction, and \(s_\ell\) contains the signed projection scores of all retained contexts onto that direction. Ignoring the numerical safeguard, the recovered axes are orthonormal because \(u_\ell^\top u_m=q_\ell^\top K_jq_m/\sqrt{\lambda_\ell\lambda_m}\). For an exact positive eigenpair, \(s_\ell=\sqrt{\lambda_\ell}q_\ell\). If \(\lambda_\ell\) is zero or numerically negligible, the corresponding axis is not a defined readout direction and should not drive a geometry label.

For the axis or antipodal diagnostic, FEGA uses the leading defined component \(s_1\). Since the sign of \(u_1\) is arbitrary, only the occupancy of both sides of the axis matters: the axis-balance score is the smaller fraction of retained contexts with positive and negative entries in \(s_1\). This tests whether an apparently one-dimensional cloud is better interpreted as an undirected axis than as one directed ray.

\section{Centered Residual Spectrum}
\label{app:centered_residual_spectrum}

The centered residual spectrum asks whether there is structured context-dependent variation after removing the average normalized effect direction. Fix a feature \(j\), and let \(V_j\) contain its \(n_j\) normalized logit-effect directions as rows. With \(H_j=I-\frac{1}{n_j}\mathbf 1\mathbf 1^\top\), row centering gives \(H_jV_j\), and since \(K_j=V_jV_j^\top\), the centered kernel is:

\[
K_{j,\mathrm{ctr}}= H_jK_jH_j .
\]

Thus,  \(K_{j,\mathrm{ctr}}\) is the Gram matrix of the centered directions. Since the trace of a Gram matrix equals total squared row norm, \(\operatorname{tr}(K_j)=\|V_j\|_F^2\) and \(\operatorname{tr}(K_{j,\mathrm{ctr}})=\|H_jV_j\|_F^2\). The residual-energy statistic is, therefore,

\[
E_{\mathrm{res}}(j)=
\frac{\operatorname{tr}(K_{j,\mathrm{ctr}})}
{\operatorname{tr}(K_j)+\epsilon},
\]

the fraction of normalized directional energy left after subtracting the mean direction.

This diagnostic asks a different question from directed-ray concentration. A feature may have a reliable average removal direction while still varying systematically across contexts; \(E_{\mathrm{res}}(j)\) measures how much such variation remains. Let \(\eta_1\geq\eta_2\geq\cdots\) be the eigenvalues of \(K_{j,\mathrm{ctr}}\). The centered-residual sufficiency score is:

\[
S_{\mathrm{res}}^{(k)}(j)=
\frac{\sum_{\ell=1}^k\eta_\ell}
{\sum_\ell\eta_\ell+\epsilon}.
\]

This score asks whether remaining deviations are low-dimensional. It is interpreted with \(E_{\mathrm{res}}(j)\): high \(S_{\mathrm{res}}^{(k)}(j)\) is meaningful only when the centered residual energy is large enough to matter, and weak when the deviations around the mean are numerically small.

\section{Directional Mixtures and Model Selection}
\label{app:vmf_model_selection}

FEGA fits the directional-mixture candidates for every feature with at least eight retained directions, independently of the preceding ray diagnostic; the family priority is applied only after these candidates have been evaluated.

For feature \(j\), the fitted data are \(y_i=v_j^{(i)}\in\mathbb S^{|\mathcal V|-1}\). If effects are stored in pre-logit form, this diagnostic materializes the corresponding logit direction as \(y_i=W_U\delta_j^{(c_i)}/\|W_U\delta_j^{(c_i)}\|_2\), since vMF fitting requires explicit logit coordinates rather than only pairwise Gram entries.

An \(M\)-component von Mises--Fisher mixture has likelihood

\[
p_M(y_i)=
\sum_{r=1}^M
\pi_r C_{|\mathcal V|}(\kappa_r)\exp(\kappa_r\mu_r^\top y_i),
\qquad
\|\mu_r\|_2=1,\quad
\kappa_r\geq 0,\quad
\sum_{r=1}^M\pi_r=1,
\]

where \(\mu_r\) is the prototype direction of mode \(r\), \(\kappa_r\) measures its concentration, and \(\pi_r\) is its fitted probability mass. These weights describe how the fitted model distributes the effect cloud across its directional modes: a context may contribute to several modes to different degrees and is assigned to its best-matching mode only when a mode-specific summary is required \citep{JMLR:v6:banerjee05a}. The normalizer \(C_d(\kappa_r)\) is retained because BIC compares full likelihoods; changing \(\kappa_r\) changes not only cosine alignment but also how probability mass is distributed on the sphere.

Writing $d=|\mathcal V|$, the normalizer is $C_d(\kappa)=\frac{\kappa^{d/2-1}}{(2\pi)^{d/2}I_{d/2-1}(\kappa)},~\kappa>0,$ with the uniform-sphere limit $C_d(0)=\frac{\Gamma(d/2)}{2\pi^{d/2}}.$ We evaluate $\log C_d(\kappa)$ in float64 using an exponentially scaled modified Bessel function. If this evaluation underflows or is otherwise non-finite, we evaluate the equivalent mode-centered spherical integral by adaptive quadrature and retain it only when the estimated relative integration error is at most $10^{-11}$ \citep{DBLP:journals/corr/abs-1907-10121, DBLP:journals/amai/Lozier03}. The implementation was checked for $2\leq d\leq256{,}000$ over the reachable concentration range, extending to $1.28\times10^{15}$ at $d=256{,}000$. The largest observed error in the cancellation-safe quantity $\log C_d(\kappa)+\kappa$ was $2.33\times10^{-10}$, below the declared $10^{-8}$ absolute bound.

For each eligible cloud, we fit every $M\in\{1,2,3,4\}$ by soft EM~\citep{JMLR:v6:banerjee05a} with k-means++ initialization~\citep{DBLP:conf/soda/ArthurV07}. Each mode count receives four deterministic starts, each limited to 200 iterations. A start stops when the summed squared change in its prototype directions is no greater than $10^{-6}$ times the mean coordinate-wise variance of the fitted unit directions. For a fixed $M$, we retain the start with the greatest finite full log likelihood; because replacement requires strict improvement, an exact tie retains the earlier start. A failed start does not suppress the remaining starts, and a mode count is unavailable only when none produces a finite full likelihood.

Candidate mode counts are selected by BIC \citep{schwarz1978estimating},

\[
\mathrm{BIC}_M = -2\log L_M+\nu_M\log n_j,
\qquad
\nu_M=M(d-1)+M+(M-1).
\]

The three terms in \(\nu_M\) count unit mean directions, concentration parameters, and free mixture weights. Beginning with the smallest finite mode count, a larger $M$ is selected only when its BIC is more than $10^{-9}$ lower; otherwise, the smaller model is retained.

BIC proposes a statistical fit, but FEGA accepts a multi-mode geometry only after additional gates. The selected model must have \(M>1\), each mode must have fitted mass $\pi_r \geq 0.10$, and its within-mode concentration is computed only when at least two contexts have that mode as their best match, within-mode directed-ray concentrations must be defined and sufficiently coherent, and clustering must improve over the pooled cloud:

\[
\Delta_{\mathrm{mix}}(j) =
\sum_{r=1}^M \pi_r C_{\mathrm{ray},r}(j) -
C_{\mathrm{ray}}(j).
\]

The selected fit is accepted as a directional mixture only when $\Delta_{\mathrm{mix}}(j)\geq0.10$, every mode has fitted mass $\pi_r\geq0.10$, every $\kappa_r$ is finite, every within-mode directed-ray concentration is defined and at least $0.70$, and assignment stability is at least $0.80$. A within-mode concentration is computed only when at least two contexts have that mode as their best match; this hard-assignment requirement is distinct from the fitted mass $\pi_r$.

For a selected $M>1$, assignment stability is evaluated on eight deterministic subsets drawn without replacement, each containing $\min\!\{n_j,\max\!\left[M,\left\lceil0.8n_j\right\rceil\right]\}$ contexts. Each subset is refitted at the already selected mode count, and its assignments are compared with the corresponding full-fit assignments using the adjusted Rand index. Assignment stability is the mean of the eight scores only when every refit succeeds and remains finite; otherwise, it is unavailable. For $M=1$, assignment stability is not applicable.

FEGA reports the selected $M$, $\Delta_{\mathrm{mix}}(j)$, the smallest fitted mode mass, the weakest within-mode concentration, assignment stability, and the smallest fitted $\kappa_r$. A rejected or unavailable fit provides no accepted multi-mode evidence, while selection of $M=1$ should not be read as proof that the cloud is truly unimodal. An accepted mixture indicates multiple directional regimes in the sampled effect cloud, not by itself multiple semantic meanings of the feature.

\section{Stability Protocols}
\label{app:fega_stability_protocols}

\begin{table}[t]
\centering
\footnotesize
\setlength{\tabcolsep}{3.5pt}
\renewcommand{\arraystretch}{0.95}
\begin{tabular}{@{}P{0.29\linewidth}P{0.17\linewidth}P{0.48\linewidth}@{}}
\arrayrulecolor{black}
\hline
Evidence gate & Threshold & Interpretation \\
\hline
\arrayrulecolor{gray!25}
Minimum retained contexts & $8$ & Below this, valid nonzero effects are too few for a geometry-family claim. \\
\hline
Maximum zero-effect filter fraction & $0.30$ & A retained coherent cloud is weak evidence if many candidate contexts showed no measurable effect. \\
\hline
Magnitude-instability flag threshold & $1.00$ & Large relative strength variation is reported as a flag, not as its own direction family. \\
\hline
90th percentile of the maximum principal angle & $30^\circ$, $30^\circ$, $35^\circ$ & The selected
  1D, 2D, or higher-D subspace should reappear across the resampled context subsets. \\
\arrayrulecolor{black}
\hline
\end{tabular}
\caption{\textbf{FEGA evidence gates.} These gates check whether a sampled feature-effect cloud has enough usable evidence for a geometry claim. They are conservative because normalization can make small or rare effects appear geometrically organized. The angle row gives the 1D, 2D, and higher-D resampling thresholds.}
\label{tab:fega_evidence_gates}
\end{table}

\begin{table}[t]
\centering
\footnotesize
\setlength{\tabcolsep}{3.5pt}
\renewcommand{\arraystretch}{0.95}
\begin{tabular}{@{}P{0.34\linewidth}P{0.14\linewidth}P{0.46\linewidth}@{}}
\arrayrulecolor{black}
\hline
Directional-family gate & Threshold & Interpretation \\
\hline
\arrayrulecolor{gray!25}
Directed-ray concentration & $0.80$ & Valid effects should repeatedly align with one directed readout direction. \\
\hline
Single-axis sufficiency, $S_{\mathrm{span}}^{(1)}$ & $0.80$ & The cloud should be nearly one-dimensional before deciding between ray and sign-split axis. \\
\hline
Ray participation-rank boundary, $d=2$ & $1.45$ & Flags ray-like cases whose spectrum is close to a low-dimensional span regime. \\
\hline
Minimum axis-balance mass & $0.15$ & Both sides of the leading axis need enough contexts to support an axis label. \\
\hline
Minimum mixture-gain improvement & $0.10$ & Clustering should make modes more ray-like than the pooled cloud. \\
\hline
Minimum mode mass & $0.10$ & Accepted modes should not be tiny leftover clusters. \\
\hline
Minimum within-mode ray concentration & $0.70$ & Each accepted mode should be internally coherent, not merely separated. \\
\hline
Minimum assignment stability & $0.80$ & Directional modes should persist under resampling, not reflect a fitting accident. \\
\arrayrulecolor{black}
\hline
\end{tabular}
\caption{\textbf{Directional-family gates.} These thresholds decide whether low or ambiguous global ray concentration is better explained by a directed ray, an unsigned axis, or several coherent directional modes}
\label{tab:fega_directional_family_gates}
\end{table}

FEGA first selects a geometry family from the complete effect cloud and, where applicable, the smallest supported dimension. It then evaluates only the stability evidence associated with that selected result as shown in Table~\ref{tab:fega_selected_family_stability}.

\begin{table}[t]
\centering
\footnotesize
\begin{tabular}{@{}P{0.27\linewidth}P{0.67\linewidth}@{}}
\arrayrulecolor{black}
\hline
Selected result & Retained stability evidence \\
\hline
\arrayrulecolor{gray!25}
Directed ray & $C_{\mathrm{ray}}$ interval and family-local leave-out and sample-size checks. \\
\hline
Axis or antipodal & $C_{\mathrm{ray}}$ interval, raw one-dimensional angle, and family-local leave-out and sample-size checks. \\
\hline
Directional mixture & Assignment stability from the selected standalone vMF fit. \\
\hline
Global span at $k$ & Raw angle at $k$ and family-local checks for strict dimensions through $k$. \\
\hline
Centered residual at $k$ & Centered-residual angle at $k$ and family-local checks for strict dimensions through $k$. \\
\hline
Fallback or terminal result & No stability protocol; reported as not evaluated. \\
\arrayrulecolor{black}
\hline
\end{tabular}
\caption{\textbf{Selected-family stability evidence.} Stability qualifies the result selected from the full cloud without reopening the family or dimension decision.}
\label{tab:fega_selected_family_stability}
\end{table}

For small or structured context sets, FEGA also uses leave-out sensitivity. Leave-one-out or leave-group-out checks remove a context, entity group, or template group and recompute the evidence required by the selected family. Rather than reclassifying each subset, FEGA asks whether the evidence continues to support that family and, for dimensioned families, the same smallest $k$. Failures weaken the confidence or add an instability flag; they do not replace the full-sample family or dimension.

For a selected axis, global span, or centered residual family, subspace stability compares the full-sample subspace with resampled subspaces only at the retained dimension: $k=1$ for an axis and $k=k^*$ for a span or residual result. Let \(U\) and \(W\) be bases for the two subspaces, orthonormal in the geometry used to store the effects: Euclidean for logit directions, or \(G\)-orthonormal for pre-logit coordinates. The singular values of \(U^\top W\) in logit coordinates, or \(U^\top G W\) in pre-logit coordinates, give the principal angles

\[
\theta_\ell=
\arccos\!\left(\operatorname{clip}(s_\ell,-1,1)\right),
\qquad
\ell=1,\ldots,k .
\]

For $n_j\geq32$, FEGA draws 20 deterministic subsets without replacement, each containing $\lceil0.75n_j\rceil$ contexts. For each subset, it converts the $k$ principal angles to degrees and retains the largest one; the reported statistic is the linearly interpolated 90th percentile of these 20 maxima. The selected subspace passes this check when the statistic is at most $30^\circ$ for $k\in\{1,2\}$ and at most $35^\circ$ for $k>2$. Each basis is obtained by eigendecomposing a symmetrized induced row kernel. An eigenvalue below $-10^{-5}$ makes the comparison unavailable; otherwise, negative eigenvalues are treated as numerical roundoff, and only eigenvalues strictly greater than $10^{-8}$ are retained. If the full cloud or any required subset has rank below $k$, the principal-angle evidence is likewise reported as unavailable. Directional mixtures instead use the fixed-mode adjusted-Rand protocol in Appendix~\ref{app:vmf_model_selection}; they do not use the scalar intervals or principal-angle calculation described here.

The evidence threshold is eight retained contexts. If $n_j<8$, FEGA reports insufficient effect evidence and does not perform stability qualification. For strict non-mixture selections, complete evidence is exploratory when $8\leq n_j<32$ and accepted when $32\leq n_j\leq64$. An accepted directional mixture instead inherits the assignment-stability evidence of its standalone fit. Fallback and terminal outcomes retain their point-state interpretation and are marked as not evaluated.

\begin{table}[t]
\centering
\footnotesize
\setlength{\tabcolsep}{3.5pt}
\renewcommand{\arraystretch}{0.95}
\begin{tabular}{@{}P{0.38\linewidth}P{0.18\linewidth}P{0.38\linewidth}@{}}
\arrayrulecolor{black}
\hline
Span or residual gate & Threshold & Interpretation \\
\hline
\arrayrulecolor{gray!25}
Span sufficiency, $S_{\mathrm{span}}^{(k)}$ & $0.90$ & The selected $k$-span should explain most normalized directional energy. \\
\hline
Component-use share, $U_{\mathrm{span}}^{(k)}$, $k=(2,3,4,8)$ & $\begin{array}{@{}c@{}}(0.08,0.05,\\ 0.03,0.01)\end{array}$ & The selected component should carry nontrivial energy, not merely complete the span. \\
\hline
Span participation rank, $r_{\mathrm{span,PR}}$, $k=(2,3,4,8)$ & $\begin{array}{@{}c@{}}(1.60,2.30,\\ 3.00,5.00)\end{array}$ & The spectrum should meaningfully use about $k$ dimensions; this is a lower-bound gate. \\
\hline
Post-$k$ span drop, $D_{\mathrm{span}}^{(k)}$ & $0.60$ & Lower is cleaner: energy should drop after $k$ rather than continue as a tail. \\
\hline
Centered residual energy & $0.10$ & Residual spectra matter only when enough variation remains after removing the mean direction. \\
\hline
Residual sufficiency, $S_{\mathrm{res}}^{(k)}$ & $0.80$ & The selected residual span should explain most centered residual directional energy. \\
\hline
Centered participation rank, $r_{\mathrm{ctr,PR}}$, $k=(2,3,4)$ & $\begin{array}{@{}c@{}}(1.50,2.20,\\ 2.90)\end{array}$ & Residual structure should meaningfully use the reported residual dimension. \\
\hline
Long-tail flag, $L_{\mathrm{tail}}$ & $1.50$ & A long-tail flag is added when entropy rank greatly exceeds participation rank. \\
\arrayrulecolor{black}
\hline
\end{tabular}
\caption{\textbf{FEGA span and residual gates.} These gates test whether a low-dimensional span is sufficient, whether its selected components are actually used, and whether the spectrum has a clean stopping point. Residual gates apply only after enough centered variation remains.}
\label{tab:fega_threshold_profiles}
\end{table}

\section{Geometry Reporting Gates}
\label{app:geometry_classifier_gates}

The geometry label is a reporting layer over pre-declared evidence gates, not a discovery procedure for internal mechanisms. Before testing a geometry family, FEGA assigns the insufficient-evidence terminal whenever the valid-effect count is unavailable or below eight, or the near-zero filter removes more than $30\%$ of the loaded contexts. Remaining clouds are tested in priority order: directed ray, axis or antipodal structure, accepted directional mixture, global low-dimensional subspace, and centered residual low-dimensional structure. This order favors specific explanations: a directed ray is also low-dimensional, but the ray label is more informative than a generic span label.

If no strict family passes, FEGA reports a nearest-family fallback only when partial evidence remains interpretable. For example, strong one-dimensional span evidence without enough signed agreement for a ray or enough balanced sign split for an axis is reported as one-dimensional diffuse evidence, rather than being forced into either family. The unresolved high-dimensional or diffuse label is itself an evidence-bearing fallback: it is assigned only when no family anchor applies and the cloud satisfies the high-dimensional/diffuse or long-tail condition. If neither a strict family nor a descriptive fallback is supported, FEGA reports undefined geometry.

\begin{table}[t]
\centering
\footnotesize
\setlength{\tabcolsep}{3.5pt}
\renewcommand{\arraystretch}{0.95}
\begin{tabular}{@{}P{0.21\linewidth}P{0.30\linewidth}P{0.22\linewidth}P{0.19\linewidth}@{}}
\arrayrulecolor{black}
\hline
Profile & Numbers & Label pressure & Trap avoided \\
\hline
\arrayrulecolor{gray!25}
Stable direction, unstable strength & Directions align, but magnitudes $(0.01,0.02,0.03,10.0)$ give large $\mathrm{CV}_{m,j}$. & Directed-family label plus magnitude-instability flag. & Reading directional stability as uniformly reliable control. \\
\hline
High sufficiency, weak second component & Spectrum $(0.91,0.08,0.01,\ldots)$ gives high $S_{\mathrm{span}}^{(2)}$ but weak $U_{\mathrm{span}}^{(2)}$. & Two dimensions suffice, but the evidence is mostly 1D. & Inferring a genuine plane from sufficiency alone. \\
\hline
Clean two-dimensional span & Spectrum $(0.50,0.45,0.03,0.02,\ldots)$ gives nontrivial $U_{\mathrm{span}}^{(2)}$ and $D_{\mathrm{span}}^{(2)}=0.03/0.45$. & The second component is substantive and followed by a drop. & Collapsing a real plane into a ray or axis. \\
\hline
Long-tailed span & Spectrum $(0.35,0.30,0.25,0.10,\ldots)$ has high small-$k$ sufficiency but $D_{\mathrm{span}}^{(2)}=0.25/0.30$. & The third component weakens a clean 2D label. & Turning a broad spectrum into a crisp low-D claim. \\
\hline
Insufficient retained evidence & Only five finite nonzero contexts remain after filtering. & Report insufficient effect evidence, not a geometry family. & Letting sample noise choose the classifier. \\
\hline
Many zero effects & Most active contexts are removed by $\tau_0$ before normalization. & Retained directions may be coherent, but only for the nonzero subset. & Treating rare measurable effects as feature-wide consistency. \\
\arrayrulecolor{black}
\hline
\end{tabular}
\caption{\textbf{Illustrative FEGA profiles.} These examples separate span sufficiency, component use, post-$k$ spectral drop, and magnitude variation. Together, they show why a geometry label should be qualified by strength, tail, and evidence flags.}
\label{tab:fega_illustrative_profiles}
\end{table}

The threshold profile has three layers: evidence gates decide whether the cloud is reliable enough to label; directional-family gates test for one ray, one sign-split axis, or several locally coherent modes; and span or residual gates test whether a low-dimensional claim is clean rather than driven by a dominant component or a long spectral tail. For global spans, FEGA tests $k \in \{2,3,4,8\}$ in increasing order and selects the smallest supported dimension. The strict centered-residual test instead considers $k \in \{2,3,4\}$. A one-dimensional residual may still be retained as a descriptive fallback when its residual-energy and sufficiency conditions hold, but it does not carry the same evidential status. If neither a strict test nor an informative fallback is supported, FEGA does not force a low-dimensional label.

FEGA records long-tail spectra using \(L_{\mathrm{tail}}=r_{\mathrm{span,ent}}/(r_{\mathrm{span,PR}}+\epsilon)\); high values warn that many small spectral components complicate a crisp rank interpretation. Each strict full-sample label is accompanied by selected-family confidence and a separate evidence-availability status. Complete evidence with no observed crossing gives accepted confidence, or exploratory confidence when the context count is below 32. Any completed boundary crossing, principal-angle failure, or selected-dimension mismatch gives unstable confidence. If required evidence is unavailable and no instability is observed, no confidence value is assigned; observed instability may coexist with unavailable auxiliary evidence. Fallback and terminal results retain their point-state confidence and are marked as not evaluated. Long-tail spectra and magnitude heterogeneity remain separate secondary descriptors.

\section{Illustrative Geometry Profiles}
\label{app:illustrative_geometry_profiles}

This appendix collects the FEGA reporting tables used to make the diagnostic procedure interpretable. Table~\ref{tab:fega_evidence_gates} states the evidence gates that determine whether a sampled effect cloud is usable for a geometry claim. Table~\ref{tab:fega_directional_family_gates} gives the directional-family gates used to distinguish directed rays, unsigned axes, and multiple coherent modes. Table~\ref{tab:fega_threshold_profiles} gives the span and residual gates used to assess low-dimensional sufficiency, component use, spectral drops, and residual variation. Table~\ref{tab:fega_illustrative_profiles} then gives small schematic spectra and context summaries showing why FEGA reports several diagnostics rather than collapsing the effect cloud to a single scalar. These examples are illustrative, not additional experiments; they clarify how sufficiency, component use, spectral tails, residual variation, zero-effect filtering, and magnitude instability affect the final reporting label.